\documentclass[lettersize,journal]{IEEEtran}
\usepackage{cite}
\usepackage{amsmath,amssymb,amsfonts}
\usepackage{algorithmic}
\usepackage{graphicx}
\usepackage{textcomp}
\usepackage{hyperref}
\usepackage{url}
\usepackage{booktabs}
\usepackage{makecell}
\usepackage{multirow}
\usepackage{color}
\usepackage[table]{xcolor}
\usepackage{colortbl}
\usepackage{upgreek}
\usepackage{pifont}
\hypersetup{
    colorlinks=true,
    linkcolor=magenta,
    filecolor=magenta,      
    urlcolor=magenta,
}
\def\secref#1{Sec.~\ref{#1}}
\def\figref#1{Fig.~\ref{#1}}
\def\tabref#1{Tab.~\ref{#1}}
\def\eqref#1{Eq.~(\ref{#1})}

\begin{document}

\title{UniMPR: A Unified Framework for Multimodal Place Recognition with Heterogeneous Sensor Configurations}

\author{Zhangshuo Qi$^{1}$, Jingyi Xu$^{2}$, Luqi Cheng$^{1}$, Shichen Wen$^{1}$, Yiming Ma$^{3}$, Guangming Xiong$^{*}$
\thanks{This work was supported by the National Natural Science Foundation of China under Grant 52372404.}
\thanks{$^{1}$Zhangshuo Qi, Luqi Cheng, and Shichen Wen are with Beijing Institute of Technology, Beijing, 100081, China (e-mail: 3120240323@bit.edu.cn).}%
\thanks{$^{2}$Jingyi Xu is with Shanghai Jiao Tong University,
 Shanghai, 200240, China (e-mail: jingyi.xu@sjtu.edu.cn).}%
 \thanks{$^{3}$Yiming Ma is with The University of New South Wales, Sydney, NSW 2052, Australia (e-mail: yiming.ma5@student.unsw.edu.au).}%
\thanks{$^{*}$Guangming Xiong (corresponding author) is with Beijing Institute of Technology, Beijing, 100081, China (e-mail: xiongguangming@bit.edu.cn).}%
}



\maketitle

\begin{abstract}
Place recognition is a critical component of autonomous vehicles and robotics, enabling global localization in GPS-denied environments. Recent advances have spurred significant interest in multimodal place recognition (MPR), which leverages complementary strengths of multiple modalities. Despite its potential, most existing MPR methods still face three key challenges: (1) dynamically adapting to various modality inputs within a unified framework, (2) maintaining robustness with missing or degraded modalities, and (3) generalizing across diverse sensor configurations and setups. In this paper, we propose UniMPR, a unified framework for multimodal place recognition. Using only one trained model, it can seamlessly adapt to any combination of common perceptual modalities (e.g., camera, LiDAR, radar). To tackle the data heterogeneity, we unify all inputs within a polar BEV feature space. Subsequently, the polar BEVs are fed into a multi-branch network to exploit discriminative intra-model and inter-modal features from any modality combinations. To fully exploit the network's generalization capability and robustness, we construct a large-scale training set from multiple datasets and introduce an adaptive label assignment strategy for extensive pre-training. Experiments on seven datasets demonstrate that UniMPR achieves state-of-the-art performance under varying sensor configurations, modality combinations, and environmental conditions. Our code will be released at \url{https://github.com/QiZS-BIT/UniMPR}.
\end{abstract}


\begin{IEEEkeywords}
Place recognition, multimodal fusion, SLAM, autonomous driving 
\end{IEEEkeywords}

\section{Introduction}
\label{sec:introduction}
\IEEEPARstart{P}{lace} recognition is a fundamental capability that enables autonomous systems to identify previously visited places using perceptual data~\cite{arandjelovic2016netvlad, uy2018pointnetvlad, wang2025segram}. It is critical for global localization, loop closure, and robust navigation, especially in GPS-denied environments~\cite{wang2023lf, hui2025pl}. This critical role has catalyzed rapid advancements in the field, which can be broadly categorized into two paradigms: unimodal approaches that rely on a single sensing modality, and multimodal place recognition that integrates data from multiple complementary sources. However, when seeking an optimal solution in practical deployment, a subtle trade-off often emerges between the inherent adaptability of unimodal methods and the superior performance potential of multimodal fusion.

Unimodal methods generally offer a distinct advantage in terms of ``plug-and-play" utility, whether based on vision (VPR)~\cite{keetha2023anyloc, huang2025robust}, LiDAR (LPR)~\cite{ma2023cvtnet, luo2024bevplace++}, or radar (RPR)~\cite{cait2022autoplace, gadd2024open}. Their simple sensor configuration requirements afford broad adaptability across diverse application scenarios. For instance, a VPR method works on any platform equipped with a camera, while RPR merely requires distinguishing between radar types. However, this adaptability comes at the expense of performance. In complex environments, unimodal methods often suffer from inherent sensing limitations: cameras struggle with illumination changes and limited field of view, LiDAR degrades under adverse weather, and radar data is noisy.

Multimodal place recognition (MPR) has emerged to address these limitations. By leveraging the complementary strengths of multiple sensors, MPR methods demonstrate enhanced performance potential compared to their unimodal counterparts. For example, \cite{zhou2023lcpr, xu2024explicit, qi2025gspr} fuse LiDAR and camera data to leverage complementary structural and semantic information, while~\cite{qi2025lrfusionpr} integrate LiDAR with radar to improve robustness across diverse scenarios. However, existing MPR approaches suffer from severe rigidity. They are extensively optimized for specific modality combinations and sensor configurations (e.g., fusing surround-view cameras with a LiDAR of specified beam count). This leads to stringent deployment requirements, thereby constraining their adaptability. 

\begin{figure}[t]
\centerline{\includegraphics[width=\columnwidth]{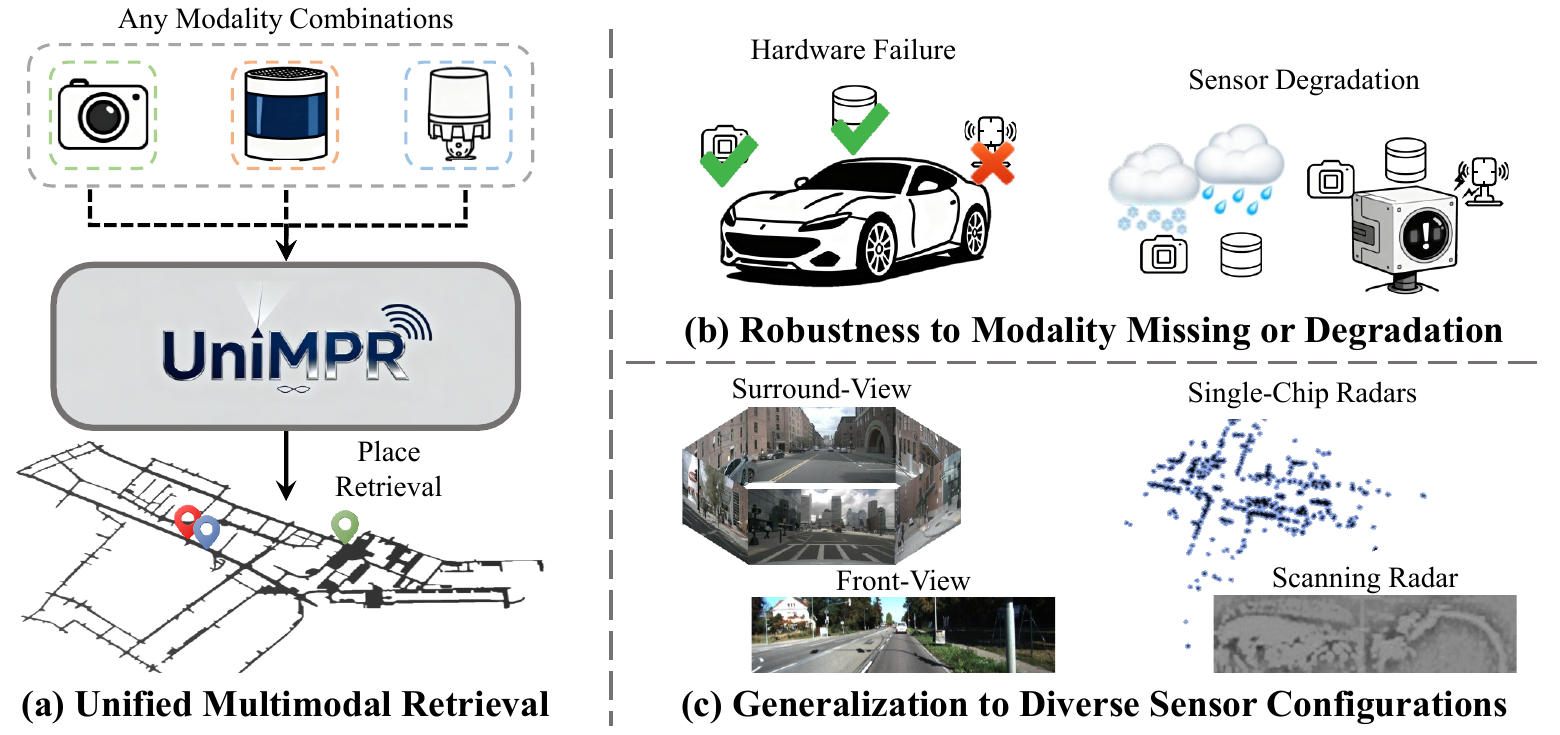}}
\caption{UniMPR unifies diverse sensor inputs for multimodal place recognition, and provides adaptability across various modality combinations and heterogeneous sensor configurations.}
\label{fig_introduction}
\vspace{-0.65cm}
\end{figure}

Specifically, the real-world adaptability of existing MPR methods is constrained by three key challenges. First, most existing MPR methods are highly specialized for a specific configuration, which limits their ability to fully exploit the potential of available hardware. For instance, when deployed on a vehicle equipped with multi-view cameras, LiDAR, and radar, an MPR system designed solely for mono camera-LiDAR fusion cannot effectively utilize the strengths offered by the multi-view cameras and radar. Second, they suffer from fragility to missing modalities or perceptual degradation. If a sensor required by the MPR system is absent on the deployment platform or becomes degraded due to environmental impact, the performance often deteriorates sharply, becoming worse than the unimodal method. Third, most MPR algorithms still exhibit limited generalization ability across varying sensor configurations. In real-world deployment, the number of cameras, types of radar, or LiDAR beam setups can differ significantly across platforms. However, how to adapt a single model to such heterogeneous setups remains insufficiently explored.

To address these challenges, we propose UniMPR, a unified MPR framework designed to bridge the gap between high accuracy and ``plug-and-play" adaptability. Unlike prior works that focus on optimizing fusion for a specific sensor suite, our core innovation is to enable a single MPR model to adapt to various modality combinations and heterogeneous sensor configurations, as shown in~\figref{fig_introduction}.

UniMPR achieves this by decoupling the data format and feature learning process from specific hardware configurations. The heterogeneous data are first unified in the polar BEV feature space. Subsequently, a multi-branch network, complemented by a learnable BEV imputation module, is introduced to extract intra-modal and inter-modal features in a manner decoupled from specific modality combinations. Ultimately, we train the network using a two-stage process on a proposed unified multimodal dataset to achieve optimal performance.

 In summary, our main contributions are as follows:
\begin{itemize}
\item We propose a MPR framework that adapts to various combinations of camera, LiDAR, and radar, enabling flexible deployment across diverse platforms.
\item We design the network and its training procedure with the specific objective of achieving resilience to modality absence or degradation.
\item We achieve strong generalization across diverse sensor configurations. The proposed method is capable of generalizing across different numbers of cameras, LiDAR beams, and radar types with a single trained model.
\item Experiments on seven datasets validate the effectiveness of our proposed method.
\end{itemize}

The remainder of this paper is organized as follows. Section II reviews the related works on unimodal and multimodal place recognition. Section III details the network architecture and training strategy. Section IV presents the experiments and ablation studies. Finally, Section V concludes the paper.

\section{Related Works}
\label{sec:related_works}
\subsection{Unimodal Methods}
Unimodal place recognition methods perform global localization using perceptual data from a single modality. As cameras, LiDAR, and radar are the most common perception sensors deployed in autonomous systems such as self-driving cars and robots, visual place recognition (VPR), LiDAR place recognition (LPR), and radar place recognition (RPR) have emerged as the three most widely studied categories.

NetVLAD~\cite{arandjelovic2016netvlad} is a pioneering work for learning-based VPR. Subsequent works have focused on extending its scope of applications. For instance, CosPlace~\cite{berton2022rethinking} introduces a training label assignment strategy for mono images, and VI\_MCPR~\cite{wang2025vi_mcpr} proposes an architecture to process multi-view images. More recently, visual foundation models have further advanced this field. AnyLoc~\cite{keetha2023anyloc} leverages foundation model to extract stable features, and MegaLoc~\cite{berton2025megaloc} reveals VPR scaling laws via large-scale fine-tuning of the foundation model.

PointNetVLAD~\cite{uy2018pointnetvlad} demonstrates the potential of LPR. LiDAR offers strong environmental robustness and captures precise geometric structures, enabling fine-grained localization. EgoNN~\cite{komorowski2021egonn} utilizes sparse convolutions to extract features from point clouds. OverlapNet~\cite{chen2022overlapnet} and OverlapTransformer~\cite{ma2022overlaptransformer} leverages range images for efficient place recognition. BEVPlace++~\cite{luo2024bevplace++} projects point clouds into BEV maps represented by point density. CVTNet~\cite{ma2023cvtnet} combines both range images and BEV representations to exploit cross-view feature correlations. SeGraM~\cite{wang2025segram} utilizes semantic information to achieve long-term loop closure detection.

RPR has received increasing attention due to its robustness to adverse weather conditions. Scanning radars, owing to their higher resolution, are first applied in RPR. OpenRadVLAD~\cite{gadd2024open} achieves place recognition using polar intensity maps and fast fourier transformation. While single-chip radar produces sparse and noisy data, some methods have successfully applied it to RPR. AutoPlace~\cite{cait2022autoplace} mitigates data sparsity by leveraging temporal information.

\subsection{Multimodal Methods}
Although unimodal place recognition methods demonstrate unique advantages, they also inherit the inherent limitations of their respective sensing modalities. Consequently, multimodal place recognition (MPR) methods have emerged, aiming to leverage the complementary strengths of multiple modalities while mitigating their individual weaknesses.

Early works primarily focused on fusing camera and LiDAR data, as they provide complementary semantic and geometric features. MinkLoc++~\cite{komorowski2021minkloc++} fuses single-view camera and LiDAR data. However, a notable limitation is its inability to effectively utilize information from multi-view camera setups. To fuse multi-view images with LiDAR, LCPR~\cite{zhou2023lcpr} explores cross-modal correlations with range images, whereas EINet~\cite{xu2024explicit} and GSPR~\cite{qi2025gspr} achieves this through explicit feature interaction. Recently, methods such as MSSPlace~\cite{melekhin2025mssplace} have explored the integration of textual data. However, ablation studies on input modalities indicate that it often fails to yield further improvements in recognition accuracy.

Furthermore, the fusion of LiDAR and radar has garnered attention in recent years. LRFusionPR~\cite{qi2025lrfusionpr} achieves LiDAR-radar fusion through cross-attention and self-distillation. Additionally, CRPlace~\cite{fu2024crplace} also explores the feasibility of multi-view camera and single-chip radar fusion.

\begin{figure*}[t]
\centerline{\includegraphics[width=\textwidth]{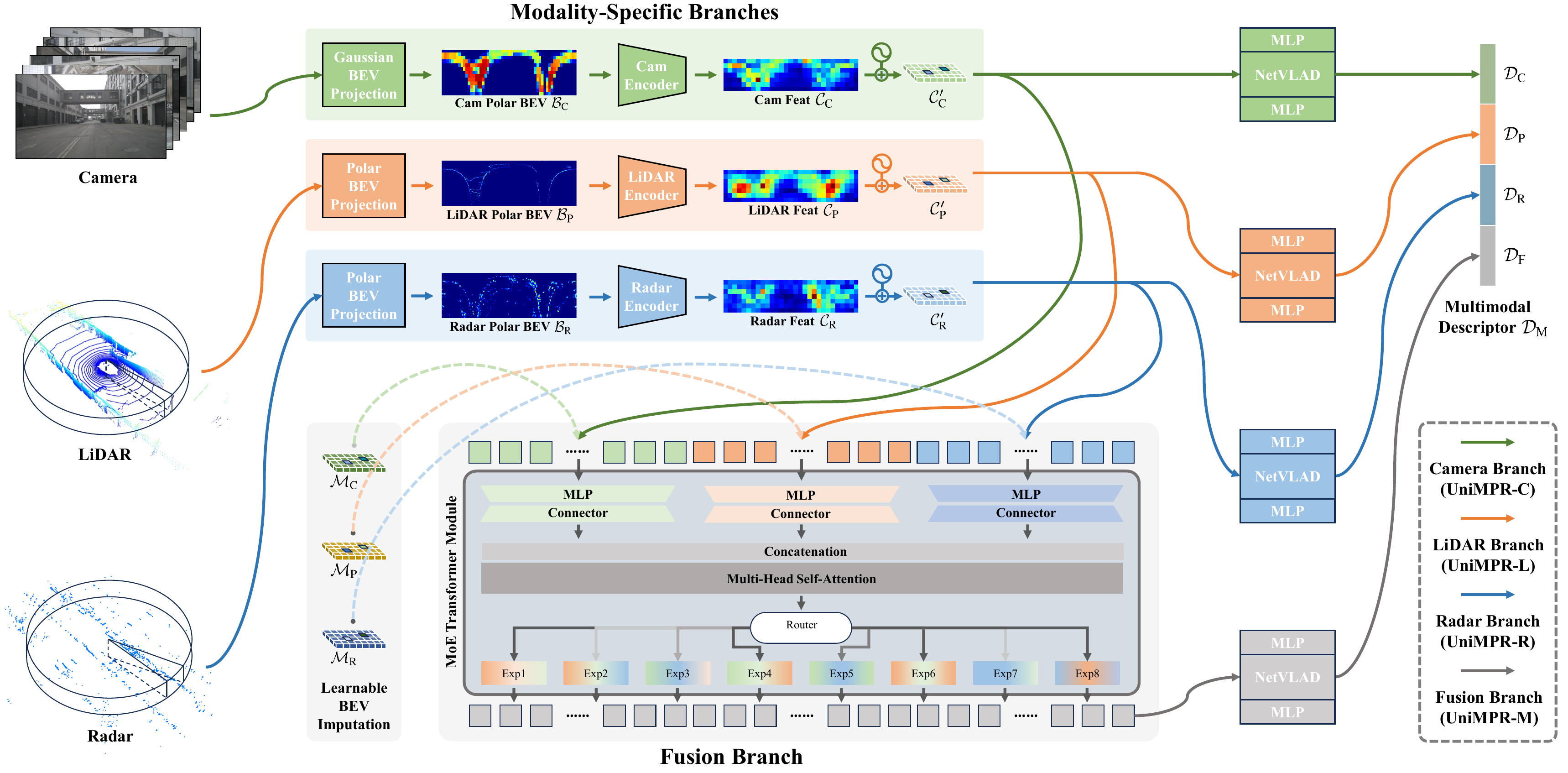}}
\caption{The overview of our proposed UniMPR. It first unifies heterogeneous data from different modalities within a polar BEV coordinate. The resulting polar BEVs are then fed into a multi-branch network. Within this network, the three modality-specific branches are designed to extract features from individual modalities, while a dedicated fusion branch learns cross-modal feature interactions. The learnable BEV imputation module is proposed to supply feature tokens for any missing modalities, thereby reducing the dependence of multimodal fusion on specific modality combinations.}
\label{fig_overall}
\vspace{-0.5cm}
\end{figure*}

Unlike existing MPR methods, our proposed framework is not reliant on specific modality combinations and can flexibly handle any subset of camera, LiDAR, and radar inputs with a unified model. Furthermore, differing from most MPR approaches that are tailored for specific sensor configurations, our method can adapt to various sensor configurations, demonstrating its robustness and strong generalization capability in different scenarios.

\section{Proposed Method}
\label{sec:proposed_method}
The overview of our proposed method is illustrated in~\figref{fig_overall}. We begin by transforming the heterogeneous data from different sensors into a unified polar BEV coordinate space. (see~\secref{sec:polar_bev_generation}). Subsequently, these polar BEV representations are fed into our devised multi-branch network to exploit inter-modal and intra-model feature correlations (see~\secref{sec:multi_branch}). Furthermore, we construct a unified multimodal dataset comprising 105k multimodal training samples and propose a two-stage training pipeline to exploit the generalization capability of the network (see~\secref{sec:dataset_and_training}).

\subsection{Polar BEV Generation}
\label{sec:polar_bev_generation}
Uniformly representing heterogeneous data from different modalities is essential for capturing cross-modal feature correlations. To ensure recognition accuracy and robustness to viewpoint changes (see~\secref{sec:polar_bev_representation}), we choose to unify the data from different modalities within the polar BEV space.
\subsubsection{LiDAR/Radar Polar BEV}
We propose using polar orthogonal projection to convert LiDAR and radar point clouds into polar BEVs. Taking the LiDAR point cloud $\mathcal{P}$ as an example, the correspondence between LiDAR point $p_{i}=\left(x_{\text{P}}, y_{\text{P}}, z_{\text{P}}\right)^{\text{T}}, p_{i}\in\mathcal{P}$ and the pixel position $\left(u_{\text{P}}, v_{\text{P}}\right)$ in polar BEV $\mathcal{B}_\text{P}$ can be represented by:
\begin{align}
    \left( \begin{array}{c} 
    u_{\text{P}} \\ 
    v_{\text{P}} 
    \end{array} \right) 
    &= \left( \begin{array}{c} 
    \frac{1}{2} \left[ 1 - \uppi^{-1} \mathrm{arctan2}\left(y_{\text{P}}, x_{\text{P}} \right) \right] w_{\text{P}} \\ 
    \left[ m^{-1} \sqrt{x_{\text{P}}^2 + y_{\text{P}}^2} \right] h_{\text{P}} 
    \end{array} \right) \label{eq:lidarp}
\end{align}
where $\left(h_{\text{P}}, w_{\text{P}}\right)$ represents the polar BEV size, and $m$ denotes the predefined maximum perception range. For generalization considerations (see~\secref{sec:polar_bev_representation}), we utilize point density as the value representation for the polar BEVs.

For the single-chip radar, we employ the same approach to project its point cloud into polar BEV $\mathcal{B}_\text{R}$. As for scanning radar, which inherently uses intensity-based polar BEV as a common representation, we retain this format.
\subsubsection{Camera Polar BEV}
\begin{figure*}[t]
\centerline{\includegraphics[width=\textwidth]{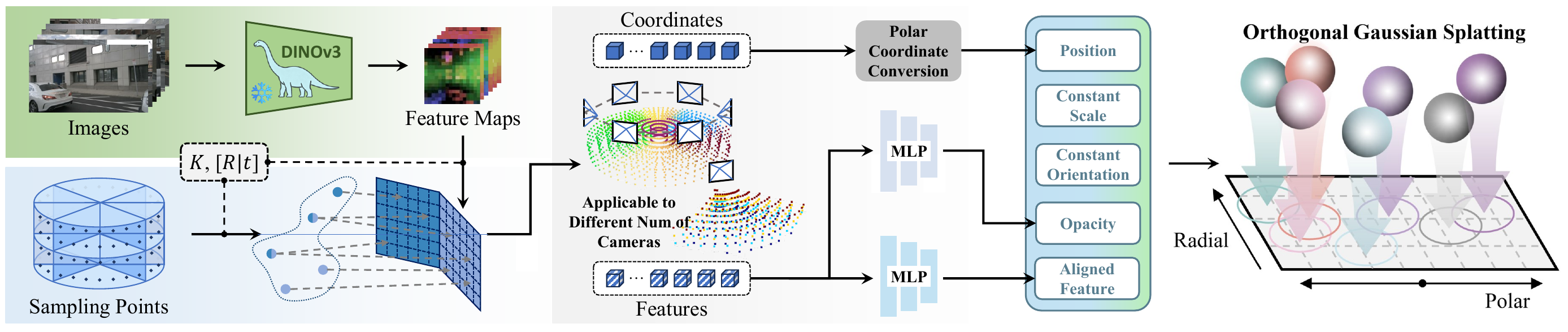}}
\caption{The Gaussian BEV Projection module. Features are lifted into 3D space via projection. We then construct 3D Gaussians from the sampled features and convert them into a polar BEV representation by orthogonal Gaussian splatting.}
\label{fig_polar_bev}
\vspace{-0.5cm}
\end{figure*}

We propose a Gaussian BEV Projection module to encode images into the polar BEV feature space. The network architecture specifically designed for the place recognition task is illustrated in~\figref{fig_polar_bev}.

To leverage the robust features extracted by foundation models for generalized MPR (as validated in~\secref{sec:image_backbone}), we employ DINOv3-S~\cite{simeoni2025dinov3} as our image backbone. Subsequently, to satisfy the low-latency demands of place recognition, we propose a parameter-free lifter that achieves 2D-to-3D feature lifting without relying on depth estimation. Our proposed BEV lifter demonstrates improved real-time performance and superior suitability for the place recognition task compared to standard BEV lifter (see~\secref{sec:gaussian_bev_projection_module}). Specifically, we define a set of points uniformly distributed in a cylindrical coordinate system as our sampling points, denoted by $\mathcal{S}$, where each point in this set can be represented as $s_{i}=[x_{i}^ \text{s},y_{i}^ \text{s},z_{i}^\text{s}], s_{i}\in\mathcal{S}$. Then their corresponding features are sampled from the feature maps, which can be defined as:
\begin{align}
    \mathbf{f}_{i}=\frac{1}{n}\sum\limits_{j\in \mathcal{N}}\mathrm{Interpolate}\left(\mathbf{F}_{j},K_{j}(R_{j}s_{i}+t_{j})\right)
\end{align}
where $\mathcal{N}$ denotes the set of images in which $s_{i}$ is visible, with $n$ representing the cardinality of $\mathcal{N}$, $\mathrm{Interpolate}()$ means the bilinear interpolation operation, $\mathbf{F}_{j}$, $K_{j}$, $R_{j}$, $t_{j}$ are the feature map, intrinsics, and extrinsics of each image, respectively.

Finally, we can obtain a set of sampling points with associated features $\mathcal{S}'$. Each point in this set can be denoted as $s'_{i}=[x_{i}^ \text{s},y_{i}^ \text{s},z_{i}^\text{s},\mathbf{f}_{i}], s'_{i}\in\mathcal{S}'$. By doing so, we lift the features into 3D space efficiently.

Subsequently, we encode features into the BEV space using Gaussian splatting orthogonal to the z-axis~\cite{kerbl20233d, lu2025toward}. Specifically, we first apply an MLP to reduce the dimensionality of the features and transform the sampling points into a cylindrical coordinate system, resulting in an intermediate representation $s''_{i}=[\rho_{i}^\text{s},\theta_{i}^ \text{s},z_{i}^\text{s},\mathbf{f}'_{i}], s''_{i}\in\mathcal{S}''$. The orthogonal Gaussian splatting operation can then be defined as follows:
\begin{align}
    \mathcal{B}_\text{C}(x)=\sum\limits_{i\in\mathcal{S}''}\mathbf{f}'_{i}\alpha_{i}\mathrm{exp}\left(-\frac{1}{2}(x-\mu_{i})^{\mathrm{T}}\Sigma^{-1}_{i}(x-\mu_{i})\right)
\end{align}
where $\mathcal{B}_\text{C}(x)$ denotes the camera polar BEV features at position $x$, $\alpha_{i}=\text{MLP}(\mathbf{f}'_{i})$ represents the predicted opacity, $\mu_{i}=[\rho_{i}^\text{s},\theta_{i}^ \text{s},z_{i}^\text{s}]$ indicates the coordinates of the Gaussian center, and $\Sigma^{-1}_{i}$ corresponds to the covariance matrix. Considering that the impact of the covariance on recognition accuracy is negligible within our framework (validated in~\secref{sec:gaussian_bev_projection_module}), we set it as a constant for real-time performance and generalization.

Through orthogonal Gaussian splatting, we spatially encode features into the polar BEV representation with learnable weighting. Furthermore, the pipeline is independent of the specific number and arrangement of cameras, thereby ensuring the network's adaptability to various sensor configurations.

\subsection{Multi-Branch Architecture}
\label{sec:multi_branch}
To process the unified polar BEV representations, we design a multi-branch network. The architecture comprises three modality-specific branches and one fusion branch, along with a learnable BEV imputation module to handle varying and potentially incomplete modality inputs.
\subsubsection{Modality-Specific Branches}
We design modality-specific branches to extract intra-modal features and modality-specific descriptors from the camera, LiDAR, and radar polar BEVs, respectively. Feature extraction is performed by encoders based on the ResNet architecture~\cite{he2016deep}. Given that the polar BEV inputs from camera, LiDAR, and radar $\{\mathcal{B}_\text{C}, \mathcal{B}_\text{P}, \mathcal{B}_\text{R}\}$ have different resolutions, we construct each modality-specific encoder using a varying number of ResBlocks. This ensures that the extracted features $\{\mathcal{C}_\text{C}, \mathcal{C}_\text{P}, \mathcal{C}_\text{R}\}$ are spatially aligned and dimensionally consistent. We then apply a learnable positional encoding to the feature maps. 

Subsequently, the position-encoded feature maps $\{\mathcal{C}'_\text{C}, \mathcal{C}'_\text{P}, \mathcal{C}'_\text{R}\}$ are fed into separate NetVLAD~\cite{arandjelovic2016netvlad} layers to extract the modality-specific descriptors ${\mathcal{D}_\text{C}, \mathcal{D}_\text{P}, \mathcal{D}_\text{R}}\in\mathbb{R}^{D}$. The adoption of NetVLAD is motivated by its effectiveness for extracting high-level abstract features, as discussed in~\secref{sec:feature_aggregator}. The modality-specific branches allow the model to fully exploit intra-modal characteristics, ensuring that performance is not solely dependent on cross-modal interactions or specific modality combinations. This enhances the network's ability to handle various modality inputs and improves its robustness to modality missing or degradation.
\subsubsection{Fusion Branch}
Meanwhile, the position-encoded feature maps are also fed into the fusion branch to fully exploit cross-modal feature correlations. We utilize a MoE Transformer structure, composed of a transformer encoder~\cite{vaswani2017attention} and a Sparse Mixture-of-Experts layer~\cite{shazeer2017outrageously}, as our feature fusion module. This architecture is designed to capture cross-modal attention relationships and to alleviate potential gradient conflicts among modalities, as motivated by~\cite{yun2024flex}. The effectiveness of this design is validated in~\secref{sec:moe_transformer}.

We first flatten the feature maps $\{\mathcal{C}'_\text{C}, \mathcal{C}'_\text{P}, \mathcal{C}'_\text{R}\}$ into modality-specific tokens. Three independent MLP connectors then project the tokens of each modality into a shared latent space, yielding representations $\{\mathcal{T}_\text{C}, \mathcal{T}_\text{P}, \mathcal{T}_\text{R}\}$ with aligned channel dimensions. These are subsequently concatenated into a unified token sequence $\mathcal{T}_\text{G}$. Since $\{\mathcal{T}_\text{C}, \mathcal{T}_\text{P}, \mathcal{T}_\text{R}\}$ are spatially aligned, each token in $\mathcal{T}_\text{G}$ contains features from all modalities within a specific spatial volume. $\mathcal{T}_\text{G}$ is then passed through a multi-head attention layer to capture global attention correlations:
\begin{align}
    \bar{\mathcal{T}_\text{G}'} = \mathrm{Attention}(\mathcal{Q}_\text{G},\mathcal{K}_\text{G},\mathcal{V}_\text{G})=\mathrm{softmax}\left(\frac{\mathcal{Q}_\text{G}\mathcal{K}_\text{G}^{\mathrm{T}}}{\sqrt{d_{\text{k}}}}\right)\mathcal{V}_\text{G}
\end{align}
where $\bar{\mathcal{T}_\text{G}'}$ represents the feature split of attention-enhanced token sequence $\mathcal{T}_\text{G}'$, $\mathcal{Q}_\text{G}$, $\mathcal{K}_\text{G}$, $\mathcal{V}_\text{G}$ are the query, key, and value splits obtained by linearly transforming the splits of $\mathcal{T}_\text{G}$.

The fused tokens $\mathcal{T}_\text{G}'$ are then fed into the MoE layer. At this stage, $E$ MLP experts are sparsely activated based on the output of the router, enabling specialized processing of different tokens. The process can be formulated as follows:
\begin{align}
    \mathcal{T}_\text{G,q}'' = \sum\limits_{e=1}^{E}\mathrm{TopK}(\text{softmax}(g(\mathcal{T}_\text{G,q}')), k)_{e} \cdot f_{e}(\mathcal{T}_\text{G,q}')
\end{align}
where $\mathcal{T}_\text{G,q}''$ means the $q$-th token in $\mathcal{T}_\text{G}''$, $g(.)$ is the gating function of the router, $f(.)$ represents the MLP expert, and $\mathrm{TopK}(.,k)$ returns a weight matrix of shape $1 \times E$, representing the activation weight for each expert. In this operation, the top-$k$ weights with the largest values are retained, while the remaining values are set to zero, thereby achieving sparse activation of experts.

\subsubsection{Learnable BEV Imputation}
We also design a learnable BEV imputation module to ensure the fusion branch remains robust to missing modalities. This module consists of learnable parameters $\{\mathcal{M}_\text{C}, \mathcal{M}_\text{P}, \mathcal{M}_\text{R}\}$ that are dimensionally equivalent to the flattened BEV feature maps $\{\mathcal{T}_\text{C}, \mathcal{T}_\text{P}, \mathcal{T}_\text{R}\}$. When a specific modality is missing, the learnable parameters are used as a substitute input into the MoE Transformer. By doing so, the impact of missing modalities can be substantially minimized (see~\secref{sec:learnable_bev_imputation}). In our framework, the modality-specific branches and the learnable BEV imputation function synergistically to handle various modality inputs. Specifically, the modality-specific branches enable adaptability during intra-modal feature extraction by selectively activating only for the provided sensors. Subsequently, the learnable BEV imputation facilitates adaptability during inter-modal feature interaction, preventing the fusion branch from overfitting to any specific modality combinations.

Finally, the fused features are fed into a NetVLAD layer to generate the fusion descriptor $\mathcal{D}_{\text{F}}$. The final multimodal descriptor $\mathcal{D}_{\text{M}}$ is obtained by concatenating $\mathcal{D}_{\text{F}}$ with the modality-specific descriptors $\mathcal{D}_{\text{C}}$, $\mathcal{D}_{\text{P}}$, and $\mathcal{D}_{\text{R}}$. When a modality is missing, its corresponding modality-specific descriptor is replaced by a zero vector.

\begin{figure}[t]
\centerline{\includegraphics[width=\columnwidth]{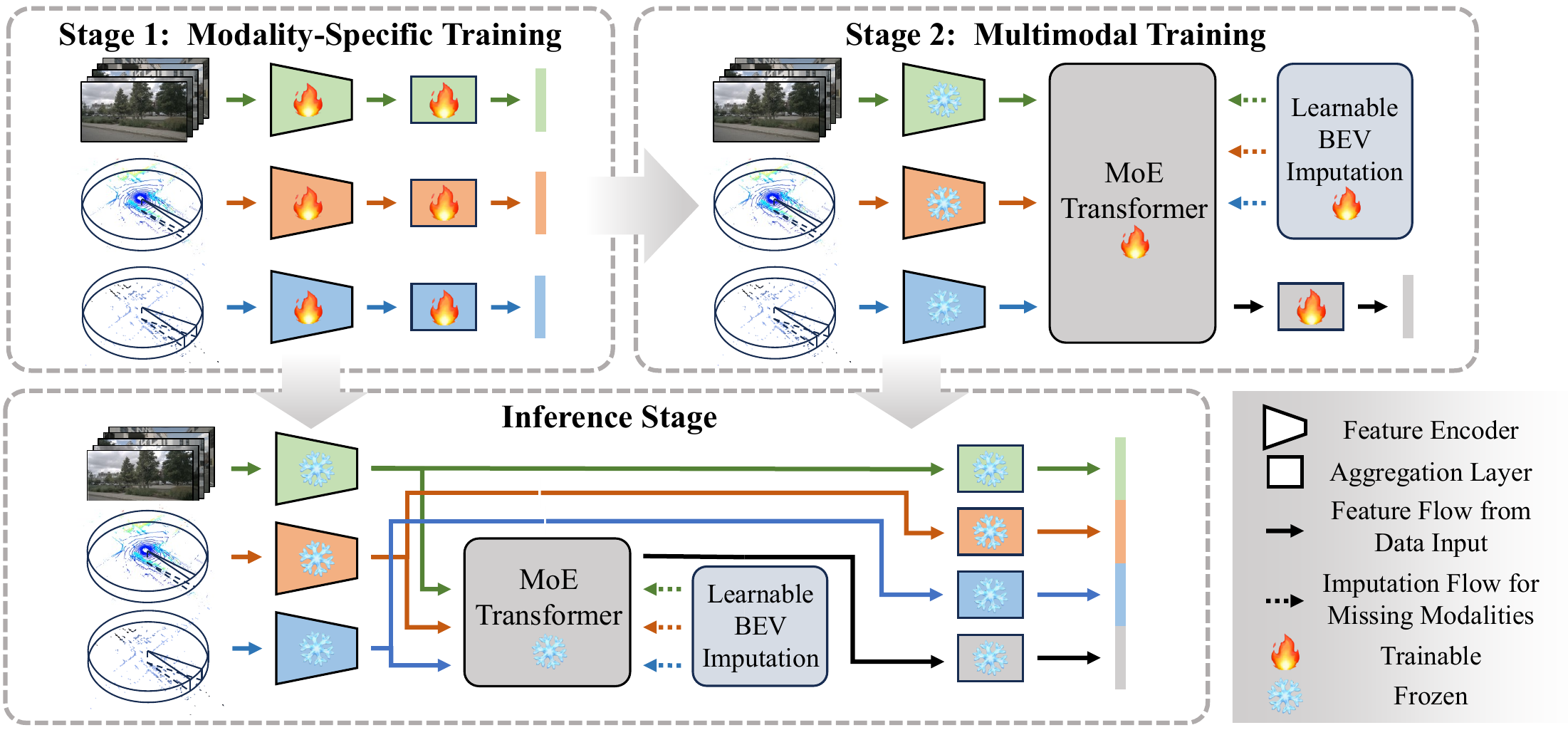}}
\caption{The proposed two-stage training pipeline. we simultaneously achieve adequate exploration of both inter-modal and intra-modal feature correlations, thereby enhancing the model's overall performance and robustness to missing modalities.}
\label{training}
\vspace{-0.4cm}
\end{figure}

\subsection{Dataset and Training Strategies}
\label{sec:dataset_and_training}
To fully exploit the potential of our network in handling various modality inputs, perceptual degradation and varying sensor configurations, we construct a unified multimodal place recognition dataset and train the model using a devised two-stage pipeline. Furthermore, to resolve label conflicts caused by varying sensor setups, we introduce an adaptive label assignment strategy.
\subsubsection{Unified Multimodal Dataset}
Inspired by the recent success of large-scale pre-training in VPR~\cite{berton2025megaloc}, we construct a unified dataset tailored for MPR. Our unified dataset is constructed primarily from four public sources: nuScenes~\cite{caesar2020nuscenes}, MulRan~\cite{kim2020mulran}, NCLT~\cite{carlevaris2016university}, and Oxford Radar~\cite{barnes2020oxford}. These datasets provide a rich variety of input modality combinations. Specifically, nuScenes and Oxford Radar offer data from all three modalities. In contrast, MulRan lacks the visual modality, while NCLT lacks radar data. Furthermore, these datasets also feature diverse sensor configurations, which is detailed in~\tabref{tab:datasets}. Through extensive pre-training on the unified dataset, our approach achieves generalized MPR with zero-shot generalization to unseen environments and sensor configurations (see~\secref{sec:zero_shot}), a capability that is challenging to achieve with limited training data (see~\secref{sec:training_strategies}).
\subsubsection{Two-Stage Training Pipeline}
Existing MPR methods exhibit poor robustness to missing modalities, largely due to the issue of \textit{modality dominance}. In end-to-end training of an MPR network, the gradient signals from highly discriminative modalities are often strong and accurate, which leads to the insufficient extraction of features from other modalities. Consequently, when operating with a missing modality, MPR methods often underperform compared to the unimodal SOTA methods that utilize only the remaining available modality (see~\secref{sec:robustness_modality}). This constitutes one of the primary reasons why existing MPR methods cannot handle various modality combinations. To address this challenge, we design a two-stage training pipeline, illustrated in~\figref{training}.

In the first stage, to ensure that each modality-specific branch can fully extract discriminative features from its corresponding modality, we independently train each branch using all of available modality-specific data. Following previous works~\cite{cait2022autoplace, zhou2023lcpr}, we employ the lazy triplet loss to supervise this training stage. Taking the camera branch as an example, the loss function is defined as:
\begin{align}
    \mathcal{L}_{\text{LT}}=\left[\beta+d(\mathcal{D}_{\text{G}}^\text{q},\mathcal{D}_{\text{G}}^{\text{pos}})-\mathop{\max}_{o}(d(\mathcal{D}_{\text{G}}^\text{q},\{\mathcal{D}_{\text{G}}^{\text{neg}}\}_{o}))\right]_+
\end{align}
where $\left [ \cdots \right ]_+$ denotes the hinge loss, $d(\cdot)$ means the Euclidean distance between a pair of descriptors, $\beta$ is the predefined margin, $o$ refers to the number of negative samples in a triplet, and $\mathcal{D}_{\text{G}}^\text{q}$, $\mathcal{D}_{\text{G}}^{\text{pos}}$, and $\mathcal{D}_{\text{G}}^{\text{neg}}$ denote the query descriptor, positive sample, and negative sample, respectively.

In the second stage, we exploit intra-modal feature correlations by training the fusion branch using temporally-aligned pairs of multimodal data. During this phase, the weights of the modality-specific branches are frozen. For the datasets with missing modalities, the fixed feature maps are substituted by parameters from the learnable BEV imputation module. The MoE Transformer then learns to exploit global context and cross-modal correlations. We also use the lazy triplet loss to supervise the training process. Additionally, a load balancing loss~\cite{he2021fastmoe} is employed to optimize the utilization of the experts, which can be defined as follows:
\begin{align}
    \mathcal{L}_{\text{LB}}=\frac{1}{E}\frac{\sum\nolimits_{e=1}^{E}(I_{e}-\bar{I})^{2}}{\bar{I}^{2}+\epsilon}+\frac{1}{E}\frac{\sum\nolimits_{e=1}^{E}(L_{e}-\bar{L})^{2}}{\bar{L}^{2}+\epsilon}
\end{align}
where $I_{e}$ denotes the sum of gating weights for the $e$-th expert over a batch, $L_{e}$ represents the sum of probabilities of the $e$-th expert being selected within a batch, $\bar{I}$ and $\bar{L}$ are the averages across all experts, and $\epsilon$ is a predefined margin.

During training, we construct a batch by sampling $B$ triplets from each dataset to ensure a balanced contribution of samples from different datasets. To prevent excessive GPU memory consumption, we employ gradient accumulation strategy, where the backward pass is computed separately for the triplets originating from each individual dataset.

This training approach enables the features of each modality to be fully exploited, thereby effectively mitigating modality dominance. Consequently, the network maintains robust performance when faced with various modality inputs or missing modalities, as shown in~\secref{sec:training_strategies}. Furthermore, it ensures full utilization of all available training data, circumventing the data wastage caused by multimodal temporal synchronization.

\begin{figure}[t]
\centerline{\includegraphics[width=\columnwidth]{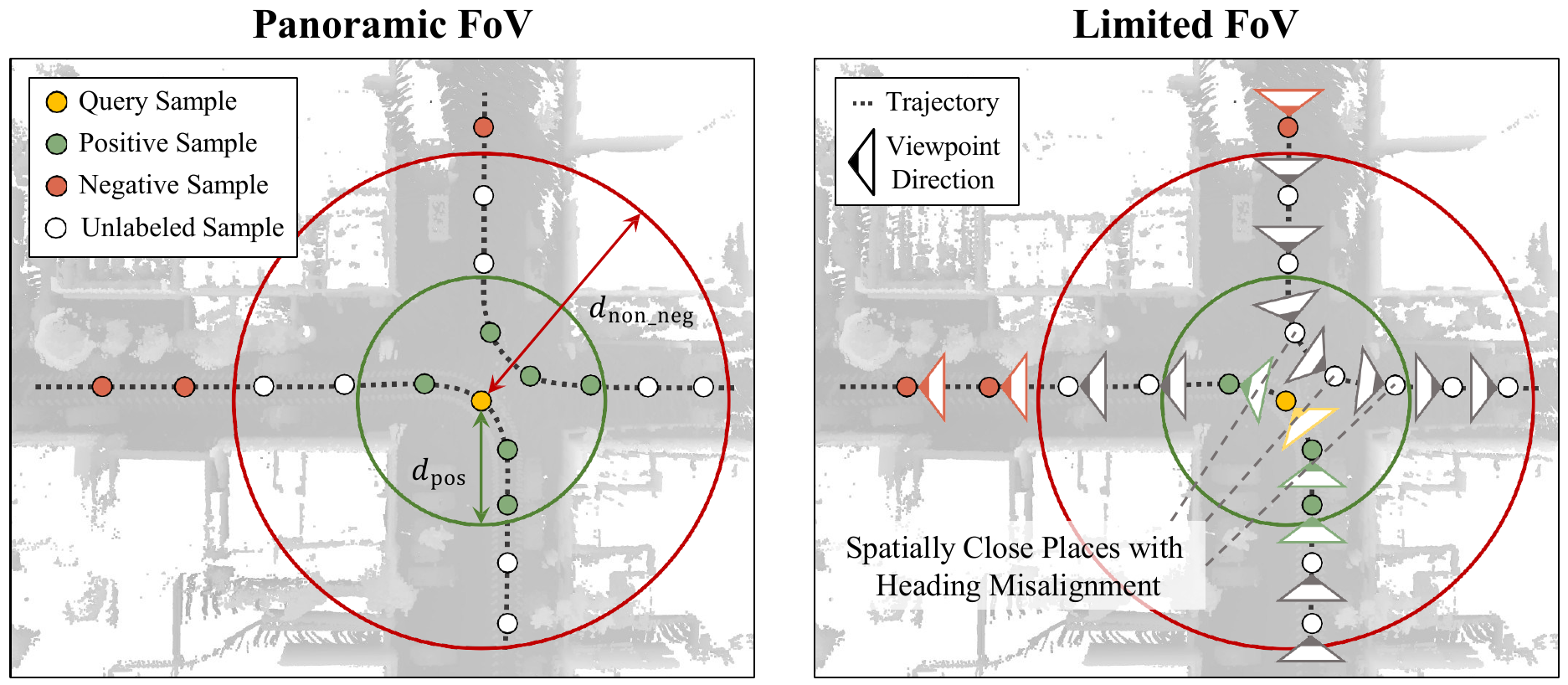}}
\caption{The proposed adaptive label assignment strategy. This strategy avoids conflicts in sample definitions by designing distinct label assignment rules for modalities with different fields of view.}
\label{label_attach}
\vspace{-0.65cm}
\end{figure}

\subsubsection{Adaptive Label Assignment Strategy}
To facilitate joint training across multiple datasets, we employ uniform spatial annotation and temporal synchronization strategies, and tailor the resampling and splitting of each dataset following previous works (detailed in~\secref{sec:dataset_setups}). However, differences in the field of view (FoV) between modalities often lead to conflicting labels for the same location, which remains an open challenge. Panoramic modalities such as LiDAR and radar rely on Euclidean distance alone to identify the positive and negative samples~\cite{luo2024bevplace++, qi2025lrfusionpr}. Conversely, limited-FoV modalities like mono cameras consider the heading angle as an additional criterion~\cite{berton2022rethinking, berton2025megaloc}. Consequently, for a place that is spatially close to the query but viewed from the opposite direction, it is considered a positive sample for panoramic modalities yet a negative sample for limited-FoV modalities, thereby creating conflicts between the modality-specific descriptors.

To address this challenge, we introduce an adaptive label assessment strategy, as shown in~\figref{label_attach}. For panoramic modalities, we retain the standard Euclidean distance-based sampling. For limited-FoV modalities, spatially close places with significant heading misalignment are treated as \textit{unlabeled} examples rather than negatives. This approach ensures label consistency under varying conditions, thereby improving the network’s performance, as shown in~\secref{sec:training_strategies}.

\section{Experiments}

\begin{table*}[ht]
  \centering
  \vspace{0.0cm}
  \setlength{\tabcolsep}{3.7pt}
  \caption{Dataset statistics and sensor configurations}
  \resizebox{\textwidth}{!}{
    \begin{tabular}{c|c|c|c|c|c|c|c}
        \toprule
        Dataset & nuScenes~\cite{caesar2020nuscenes} & NCLT~\cite{carlevaris2016university} & Oxford Radar~\cite{barnes2020oxford} & MulRan~\cite{kim2020mulran} & KITTI~\cite{geiger2013vision} & Boreas~\cite{burnett2023boreas} & Self-Collected \\ \hline
        Sensors & \makecell{Surround-view \\ cameras \\ 32-beam LiDAR \\ Single-chip radars} & \makecell{Surround-view \\ cameras \\ 32-beam LiDAR} & \makecell{Front-view camera \\ 32-beam LiDAR \\ Scanning radar} & \makecell{64-beam LiDAR \\ Scanning radar} & \makecell{Front-view cameras \\ 64-beam LiDAR} & \makecell{Front-view camera \\ 128-beam LiDAR \\ Scanning radar} & \makecell{Surround-view \\ cameras \\ 32-beam LiDAR} \\ \hline
        Timespan & Several months & Across 1 year & Several days & Several months & Single day & Across 1 year & Several months\\ \hline
        Condition & Day/Night/Rain & Day & Day/Rain & Day & Day & Day/Night/Rain/Snow & Day\\
        \bottomrule
    \end{tabular}
  }
  \label{tab:datasets}
  \vspace{-0.3cm}
\end{table*}

\begin{table*}[ht]
  \centering
  	\vspace{0.0cm}
  	\setlength{\tabcolsep}{4.8pt}
    \setlength{\abovecaptionskip}{0.15cm}
    \caption{Performance on the nuScenes dataset}
    \resizebox{\textwidth}{!}{
    \begin{tabular}{c|c|cccc|cccc|cccc}
      \toprule
      \multirow{2}{*}{Methods} & \multirow{2}{*}{Modality} &\multicolumn{4}{c|}{BS split}&\multicolumn{4}{c|}{SON split}&\multicolumn{4}{c}{SQ split} \\ \cline{3-14}
      ~ & ~ & AR@1 & AR@5 & AR@10 & max $F_{1}$ & AR@1 & AR@5 & AR@10  & max $F_{1}$ & AR@1 & AR@5 & AR@10  & max $F_{1}$ \\ \hline
      AnyLoc~\cite{keetha2023anyloc} & C & 80.55 & 89.80 & 90.85 & 0.9257 & 76.29 & 83.02 & 85.39 & 0.9432 & 57.11 & 65.22 & 68.51 & 0.8166 \\
      MegaLoc~\cite{berton2025megaloc} & C & 86.39 & 90.56 & 90.91 & 0.9711 & 80.88 & 84.31 & 84.99 & 0.9690 & 59.58 & 64.28 & 66.51 & 0.8993 \\
      EgoNN~\cite{komorowski2021egonn} & L & 89.48 & 93.26 & 94.83 & 0.9588 & 88.15 & 94.09 & 95.28 & 0.9451 & 88.84 & 94.24 & 96.12 & 0.9513 \\
      BEVPlace++~\cite{luo2024bevplace++} & L & 92.53 & 97.10 & 98.02 & 0.9620 & 93.95 & 96.74 & 97.55 & 0.9739 & 95.06 & 98.12 & 98.94 & 0.9776 \\
      AutoPlace~\cite{cait2022autoplace} & R & 78.59 & 82.92 & 83.68 & 0.9453 & 72.75 & 75.06 & 76.76 & 0.8851 & 64.12 & 66.12 & 68.00 & 0.9411 \\
      LCPR~\cite{zhou2023lcpr} & L+C & 83.06 & 90.91 & 93.72 & 0.9102 & 80.81 & 88.82 & 91.13 & 0.9020 & 61.34 & 75.56 & 80.14 & 0.8112 \\
      LRFusionPR~\cite{qi2025lrfusionpr} & L+R & \underline{97.75} & \underline{99.51} & \underline{99.73} & \underline{0.9892} & \underline{95.92} & \underline{98.68} & \underline{99.05} & \underline{0.9808} & \underline{97.65} & \underline{99.65}  & \underline{99.76} & \underline{0.9899} \\
      CRPlace~\cite{fu2024crplace} & C+R & 91.2 & 92.6 & 93.3 & 0.96 & - & - & - & - & - & - & - & - \\
      \rowcolor{gray!25}
      UniMPR (ours) & L+C+R & \textbf{99.13} & \textbf{99.86} & \textbf{99.92} & \textbf{0.9959} & \textbf{98.40} & \textbf{99.73} & \textbf{99.90} & \textbf{0.9931} & \textbf{99.65} & \textbf{99.88}  & \textbf{100.0} & \textbf{0.9982} \\
      \bottomrule
    \multicolumn{14}{p{0.9\linewidth}}{C: Camera, L: LiDAR, R: Single-Chip Radar. The best and secondary results are highlighted in \textbf{bold black} and \underline{underline} respectively.}\\
    \multicolumn{14}{p{0.9\linewidth}}{Since the code for CRPlace is not publicly available and it employs the same benchmark, results are reported as in the original paper.}\\
    \end{tabular}
    }
    \label{tab:nuscenes}
    \vspace{-0.65cm}
\end{table*}

\begin{table}[t]
  \centering
  \vspace{0.0cm}
  \setlength{\tabcolsep}{2.4pt}
  \caption{Performance on the Oxford Radar dataset}
  \resizebox{\columnwidth}{!}{
    \begin{tabular}{c|c|ccccc}
        \toprule
        \multirow{2}{*}{Methods} & \multirow{2}{*}{Modality} &  \multicolumn{5}{c}{\makecell{Query: 2019-01-18-14-46-59 \\ Database: 2019-01-10-14-50-05}}\\ \cline{3-7}
        ~ & ~ & AR@1 & AR@5 & AR@10 & AR@20 & max $F_{1}$ \\ \hline
        AnyLoc~\cite{keetha2023anyloc} & C & 98.46 & 99.50 & 99.66 & 99.70 & 0.9931 \\
        MegaLoc~\cite{berton2025megaloc} & C & 99.13 & 99.66 & 99.70 & 99.80 & 0.9959 \\
        EgoNN~\cite{komorowski2021egonn} & L & 97.98 & 99.17 & 99.46 & 99.66 & 0.9911 \\
        BEVPlace++~\cite{luo2024bevplace++} & L & 99.18 & \underline{99.67} & \underline{99.77} & \underline{99.81} & 0.9966 \\
        OpenRadVLAD~\cite{gadd2024open} & R & 90.97 & 96.22 & 97.51 & 98.62 & 0.9559 \\
        LRFusionPR~\cite{qi2025lrfusionpr} & L+R & \underline{99.31} & 99.60 & 99.63 & 99.70 & \underline{0.9981} \\
        \rowcolor{gray!25}
        UniMPR (ours) & L+C+R & \textbf{99.84} & \textbf{99.99} & \textbf{100.0} & \textbf{100.0} & \textbf{0.9992} \\
        \bottomrule
    \end{tabular}
  }
  \label{tab:oxford_radar}
  \vspace{-0.5cm}
\end{table}

\subsection{Implementation details}
\subsubsection{Dataset Setups}
\label{sec:dataset_setups}
We train our network on the unified multimodal dataset (see~\secref{sec:dataset_and_training}) and evaluate it on seven distinct datasets. The details of these datasets are summarized in~\tabref{tab:datasets}.

For the nuScenes dataset, we follow the data organization protocol in LRFusionPR~\cite{qi2025lrfusionpr}. The training set combines nuScenes-SHV and a subset of nuScenes-BS, while evaluation uses the remaining nuScenes-BS, nuScenes-SON, and nuScenes-SQ splits. For the division of database and query sets on the nuScenes-SON split, we introduce some adjustments. Specifically, only data collected during the first 15 days is used as the database, in order to evaluate the localization accuracy under greater temporal variation.

The data organization for the MulRan dataset also follows the protocol of LRFusionPR. We spatially divide the longest split, Sejong, into training and test sets. In addition to evaluating on Sejong, the MulRan-DCC sequence serves as additional unseen data for testing zero-shot generalization capability.

For the NCLT dataset, the partitioning follows RING\#~\cite{lu2025ring}, BEVPlace++~\cite{luo2024bevplace++}, and CVTNet~\cite{ma2023cvtnet}. We select ``2012-01-08", ``2012-02-04", ``2012-03-17", and ``2012-05-26" sequences for training. During testing, ``2012-01-08" serves as the database, while ``2012-06-15", ``2012-09-28", and ``2013-02-23" are used as queries. This setup is designed to validate the method's robustness under temporal variations exceeding one year.

The processing of the Oxford Radar dataset follows RING\# and LRFusionPR. The sequences ``2019-01-11-13-24-51", ``2019-01-11-14-37-14", ``2019-01-15-13-06-37", and ``2019-01-15-13-06-37" are selected for training. For testing, ``2019-01-10-14-50-05" serves as the database. ``2019-01-18-14-46-59" and ``2019-01-16-14-15-33'' are used as the query under sunny and rainy weather, respectively.

Additionally, to rigorously evaluate the zero‑shot generalization ability of our method to unseen environments, modality combinations, and sensor configurations, we introduce the Boreas~\cite{burnett2023boreas} and KITTI~\cite{geiger2013vision} datasets, as well as a self‑collected campus‑environment dataset. These three datasets exhibit entirely different collection locations, times, and sensor configurations compared to all training sets. For the Boreas dataset, we select ``2020-12-18-13-44" as the database, ``2021-01-26-11-22" as the query under adverse weather conditions, and the three sequences ``2021-04-29-15-55", ``2021-09-14-20-00", and ``2021-11-16-14-10" as queries under long-term temporal variations. For the KITTI dataset, sequences 00, 02, and 05 are selected to perform the loop-closure detection evaluation following the configuration of OverlapTransformer~\cite{ma2022overlaptransformer}. The self-collected dataset comprises two sequences recorded on the same route with a one‑month interval, as detailed in~\secref{sec:real_world}.

\subsubsection{Configurations}
We set the LiDAR and radar polar BEV sizes to $(900, 200)$ and $(225, 50)$, respectively. Images are resized to $224 \times 224$ for training and $336 \times 336$ for inference. The camera polar BEV generated via Gaussian splatting has size of $(57, 13)$. The dimension of the modality-specific and fusion descriptors are set to $D=256$, resulting in a multimodal descriptor $\mathcal{D}_{\text{M}} \in \mathbb{R}^{1 \times 1024}$. For the triplet loss, we set the margin $\beta=0.5$ and the number of hard negatives $o=10$. For all datasets, the distance threshold for positive samples and non-negative samples are uniformly set to $d_{\text{pos}}=9\,\text{m}$ and $d_{\text{non\_neg}}=12\,\text{m}$, respectively. For sensors with a limited field of view, the heading difference threshold between a positive sample and a query is consistently set to $60°$. Each complete batch consists of $B=8$ triplets sampled from each dataset. In both stages of our training pipeline, the initial learning rate is set to $5 \times 10^{-5}$ and decays by $20\%$ after each epoch. For data augmentation, random translation is applied to polar BEVs, while RandAugment~\cite{cubuk2020randaugment} is used for images. In all subsequent experiments, unless otherwise specified, we define a query as successful if the distance between the candidate and the query is within $9\,\text{m}$. For datasets that are not temporally aligned, the modality with the lowest sampling frequency serves as the reference; subsequently, data from other modalities with the smallest temporal offset to the reference are aligned accordingly. We conduct training and evaluation on a system with an Intel i7-14700KF CPU and an NVIDIA GeForce RTX 4060Ti GPU.

\subsubsection{Baselines}
We select a variety of SOTA unimodal and multimodal methods as baselines. For VPR, we employ AnyLoc~\cite{keetha2023anyloc} and MegaLoc~\cite{berton2025megaloc}, which leverage foundation models and trained on a large scale, as baselines. LPR baselines include EgoNN~\cite{komorowski2021egonn}, the SOTA method on the MulRan dataset, and BEVPlace++~\cite{luo2024bevplace++}. For RPR, we select OpenRadVLAD~\cite{gadd2024open} (scanning radar) and AutoPlace~\cite{cait2022autoplace} (single-chip radar) to cover different scenarios. Regarding MPR methods, we select LCPR~\cite{zhou2023lcpr}, LRFusionPR~\cite{qi2025lrfusionpr} and CRPlace~\cite{fu2024crplace}. For unimodal baselines, we aim to verify that our method achieves comparable adaptability while demonstrating the performance advantages of MPR. For multimodal baselines, we highlight the superiority of our approach when handling various modality inputs, perceptual degradation, and diverse sensor configurations. Moreover, we show that our method achieves competitive or even better performance than the MPR baselines under their own optimized settings, even when using identical sensor inputs, as detailed in~\secref{sec:robustness_modality}. Notably, many existing MPR methods are highly optimized for specific scenarios or sensor setups. Therefore, to ensure the credibility of the experimental results, except for robustness evaluation under missing-modality conditions, we do not adapt baselines designed for specific scenarios to other incompatible settings (e.g., forcibly adapting LCPR to scenarios that cannot provide surround-view images and 32-beam LiDAR data). 
For fair comparison, all baselines are implemented based on their official open-source code and pre-trained models.

\begin{table*}[ht]
  \centering
  	\vspace{0.0cm}
  	\setlength{\tabcolsep}{4.7pt}
    \caption{Performance on the NCLT dataset}
    \resizebox{\textwidth}{!}{
    \begin{tabular}{c|c|cccc|cccc|cccc}
      \toprule
      \multirow{2}{*}{Methods} & \multirow{2}{*}{Modality} &\multicolumn{4}{c|}{\makecell{Query: 2012-06-15 \\ Database: 2012-01-08}}&\multicolumn{4}{c|}{\makecell{Query: 2012-09-28 \\ Database: 2012-01-08}}&\multicolumn{4}{c}{\makecell{Query: 2013-02-23 \\ Database: 2012-01-08}} \\ \cline{3-14}
      ~ & ~ & AR@1 & AR@5 & AR@10 & max $F_{1}$ & AR@1 & AR@5 & AR@10  & max $F_{1}$ & AR@1 & AR@5 & AR@10  & max $F_{1}$ \\ \hline
      AnyLoc~\cite{keetha2023anyloc} & C & 74.18 & 82.87 & 85.73 & 0.8718 & 66.46 & 77.10 & 81.06 & 0.8199 & 58.46 & 66.77 & 70.32 & 0.8401 \\
      MegaLoc~\cite{berton2025megaloc} & C & 81.28 & 85.79 & 87.17 & \underline{0.9412} & 78.06 & 83.71 & 85.91 & 0.9111 & 60.82 & 64.36 & 66.38 & 0.9047 \\
      EgoNN~\cite{komorowski2021egonn} & L & 76.45 & 84.01 & 86.80 & 0.8976 & 68.70 & 78.71 & 82.32 & 0.8608 & 83.62 & 89.39 & 91.51 & 0.9113 \\
      BEVPlace++~\cite{luo2024bevplace++} & L & \underline{87.64} & \underline{92.39} & \underline{95.80} & 0.9350 & \underline{89.54} & \underline{93.95} & \underline{95.54} & \underline{0.9453} & \underline{87.73} & \underline{92.39} & \underline{93.67} & \underline{0.9359} \\
      LCPR~\cite{zhou2023lcpr} & L+C &  81.64 & 87.30 & 89.16 & 0.9098 & 69.61 & 79.80 & 83.53 & 0.8275 & 79.54 & 85.31 & 87.55 & 0.8904\\
      LRFusionPR$^{*}$~\cite{qi2025lrfusionpr} & L & 77.83 & 85.16 & 88.18 & 0.8799 & 72.90 & 82.84 & 86.19 & 0.8471 & 79.09 & 87.99 & 90.55 & 0.8833\\
      \rowcolor{gray!25}
      UniMPR$^{*}$ (ours) & L+C & \textbf{99.40} & \textbf{99.81} & \textbf{99.87} & \textbf{0.9975} & \textbf{99.56} & \textbf{99.83} & \textbf{99.92} & \textbf{0.9979} & \textbf{98.36} & \textbf{99.04}  & \textbf{99.36} & \textbf{0.9931} \\
      \bottomrule
    \multicolumn{14}{p{0.9\linewidth}}{$^{*}$ MPR methods evaluated with missing input modalities.}\\
    \end{tabular}
    }
    \label{tab:nclt}
    \vspace{-0.65cm}
\end{table*}

\begin{table}[t]
  \centering
  \vspace{0.0cm}
  \setlength{\tabcolsep}{0.6pt}
  \caption{Performance on the MulRan dataset}
  \resizebox{\columnwidth}{!}{
    \begin{tabular}{c|c|ccc|ccc}
        \toprule
        \multirow{2}{*}{Methods} & \multirow{2}{*}{Modality} & \multicolumn{3}{c|}{Sejong} & \multicolumn{3}{c}{DCC}\\ \cline{3-8}
        ~ & ~ & AR@1 & AR@5 & AR@10 & AR@1 & AR@5 & AR@10 \\ \hline
        EgoNN~\cite{komorowski2021egonn} & L & 97.32 & 99.71 & 99.86 & 72.09 & 82.17 & 82.95 \\
        BEVPlace++~\cite{luo2024bevplace++} & L & 85.51 & 90.72 & 92.39 & 70.54 & 83.72 & 86.05 \\
        OpenRadVLAD~\cite{gadd2024open} & R & 94.06 & 97.75 & 98.77 & 57.36 & 74.42 & 79.07 \\
        LRFusionPR~\cite{qi2025lrfusionpr} & L+R & \textbf{98.91} & \textbf{99.93} & \underline{100.0} & \underline{75.97} & \underline{86.82} & \underline{92.25} \\
        \rowcolor{gray!25}
        UniMPR$^{*}$ (ours) & L+R & \underline{98.62} & \underline{99.78} & \textbf{100.0} & \textbf{81.40} & \textbf{98.45} & \textbf{99.22} \\
        \bottomrule
    \multicolumn{8}{p{1.0\columnwidth}}{LCPR remains incompatible due to discrepancies in LiDAR beam, even with missing camera data padded.}\\
    \end{tabular}
  }
  \label{tab:mulran}
  \vspace{-0.3cm}
\end{table}

\begin{table}[t]
  \centering
  \vspace{0.0cm}
  \setlength{\tabcolsep}{4.5pt}
  \caption{Performance on the LiDAR-excluded nuScenes dataset}
  \resizebox{\columnwidth}{!}{
    \begin{tabular}{c|c|cccc}
        \toprule
        \multirow{2}{*}{Methods} & \multirow{2}{*}{Modality} & \multicolumn{4}{c}{BS split} \\ \cline{3-6}
        ~ & ~ & AR@1 & AR@5 & AR@10 & max $F_{1}$ \\ \hline
        CRPlace~\cite{fu2024crplace} & C+R & 91.2 & 92.6 & 93.3 & 0.96 \\
        LCPR$^{*}$~\cite{zhou2023lcpr} & C & 64.20 & 76.27 & 81.93 & 0.7907\\
        LRFusionPR$^{*}$~\cite{qi2025lrfusionpr} & R & 61.44 & 74.43 & 78.73 & 0.7708 \\
        \rowcolor{gray!25}
        UniMPR$^{*}$ (ours) & C+R & \textbf{94.53} & \textbf{98.19} & \textbf{99.03} & \textbf{0.9729} \\
        \bottomrule
    \end{tabular}
  }
  \label{tab:nusc_lidar_exclude}
  \vspace{-0.5cm}
\end{table}

\begin{table}[t]
  \centering
  \vspace{0.0cm}
  \setlength{\tabcolsep}{1.6pt}
  \caption{Performance on the KITTI dataset}
  \resizebox{\columnwidth}{!}{
    \begin{tabular}{c|c|cc|cc|cc}
        \toprule
        \multirow{2}{*}{Methods} & \multirow{2}{*}{Modality} & \multicolumn{2}{c|}{00} & \multicolumn{2}{c|}{02} & \multicolumn{2}{c}{05} \\ \cline{3-8}
        ~ & ~ & AR@1 & AR@5 & AR@1 & AR@5 & AR@1 & AR@5 \\ \hline
        AnyLoc~\cite{keetha2023anyloc} & C & 88.75 & 90.45 & 75.82 & 77.91 & 80.18 & 87.89 \\
        MegaLoc~\cite{berton2025megaloc} & C & 90.00 & 91.48 & 77.61 & 79.40 & 80.00 & 82.46 \\
        EgoNN~\cite{komorowski2021egonn} & L & 89.89 & 92.95 & 85.07 & 90.15 & 75.61 & 78.77 \\
        BEVPlace++~\cite{luo2024bevplace++} & L & \underline{94.09} & \underline{96.02} & \underline{96.12} & \textbf{98.51} & \underline{81.75} & \underline{93.68} \\
        \rowcolor{gray!25}
        UniMPR (ours) & L+C & \textbf{95.80} & \textbf{96.93} & \textbf{97.31} & \underline{97.61} & \textbf{90.88} & \textbf{97.54} \\
        \bottomrule
    \end{tabular}
  }
  \label{tab:kitti}
  \vspace{-0.25cm}
\end{table}

\subsection{Performance in Full Modality Scenarios}
In this section, we conduct experiments on the nuScenes and Oxford Radar datasets to validate our method's capability to maximize the complementary strengths of multimodal data when all modalities are available, as well as its adaptability to various sensor configurations. Following~\cite{zhou2023lcpr, qi2025lrfusionpr}, we report Recall@K and the max $F_{1}$ score as our evaluation metrics.

The results in~\tabref{tab:nuscenes} and~\tabref{tab:oxford_radar}, demonstrate that our method outperforms all baselines. This performance improvement can be attributed to the exploitation of the complementary strengths inherent in heterogeneous multimodal data, regardless of their specific configurations.

The distinct sensor characteristics of the two datasets, as detailed in~\tabref{tab:datasets}, underscore this advantage. The nuScenes dataset comprises surround-view images with single-chip radar data, whereas the Oxford Radar consists of single-view images with scanning radar data. This configuration gap renders many baselines inapplicable to one of the datasets. Methods such as LCPR and CRPlace are confined to the nuScenes dataset due to their hard requirement for surround-view images, while AutoPlace and OpenRadVLAD lack general applicability because of the significant domain gap between single-chip radar and scanning radar data. Furthermore, mono image-based VPR methods like AnyLoc and MegaLoc exhibit inherent vulnerability to severe viewpoint changes, as evidenced by their performance drop on the challenging nuScenes-SQ split.

In stark contrast, our method addresses the limitations mentioned above from two key aspects. First, it achieves robust and generalizable place recognition by adapting to diverse sensor configurations. Second, unlike MPR frameworks limited to LiDAR-camera that cannot utilize radar data, or mono image-based methods that fail to harness the rich information from surround-view data, our approach provides a unified solution that fully exploits all available multimodal information.

\subsection{Robustness to Various Modality Combinations}
\label{sec:robustness_modality}
While multimodal fusion demonstrates superior performance when all modalities are available, practical deployment often faces incomplete sensor data due to hardware constraints or environmental factors. Existing MPR methods typically fail under such conditions as they are designed for fixed modality combinations. This limitation significantly restricts their real-world applicability. In this section, we conduct experiments to show that our proposed method enables robust operation with various input modality combinations with one trained model. We evaluate this capability using three datasets: NCLT (no radar), MulRan (no camera), and a modified version of nuScenes where the LiDAR modality is artificially ignored.

\begin{figure}[t]
\centerline{\includegraphics[width=\columnwidth]{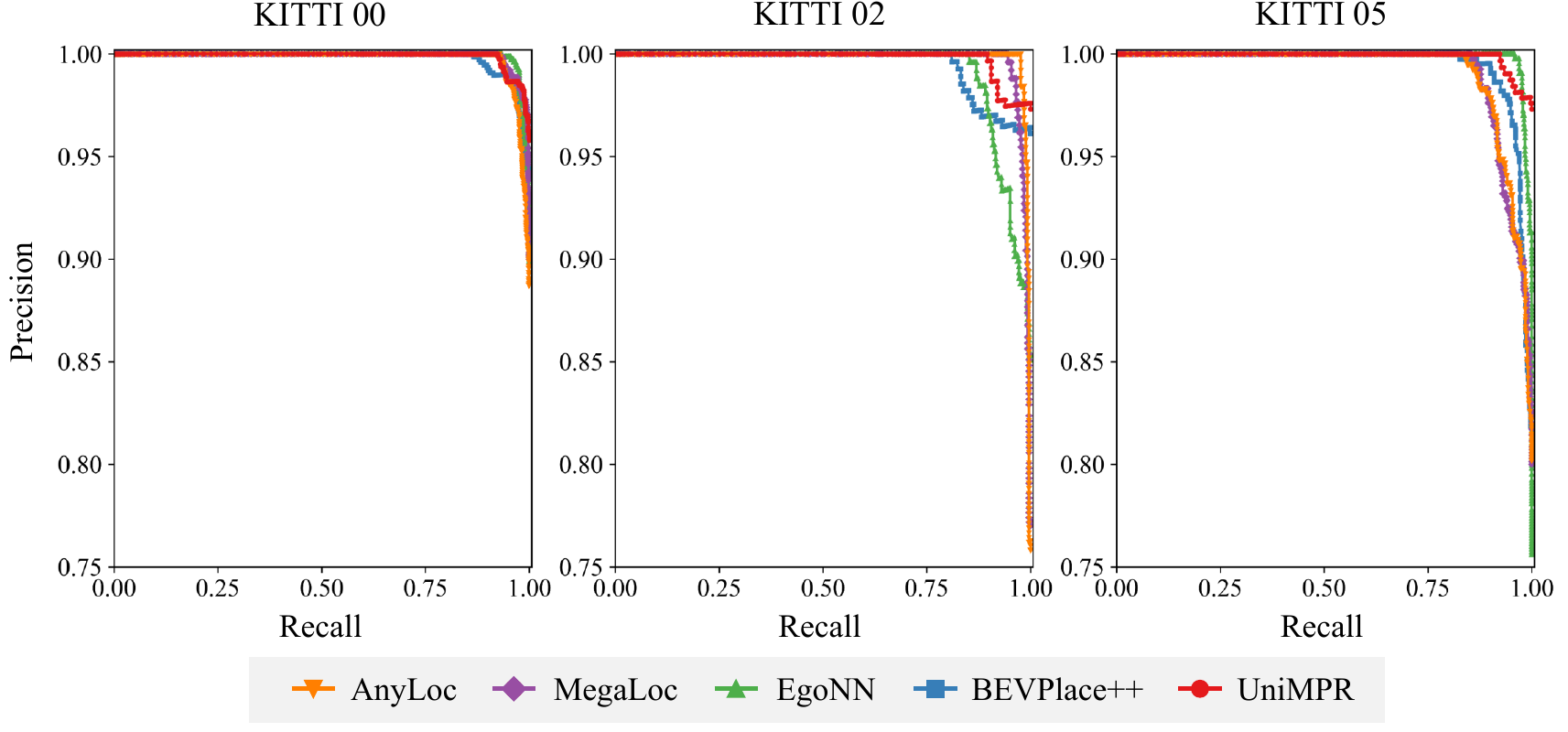}}
\caption{The precision-recall curve on the KITTI dataset.}
\label{fig:kitti_pr}
\vspace{-0.5cm}
\end{figure}

\begin{table*}[ht]
  \centering
  	\vspace{0.0cm}
  	\setlength{\tabcolsep}{4.4pt}
    \caption{Performance on the Boreas dataset}
    \resizebox{\textwidth}{!}{
    \begin{tabular}{c|c|cccc|cccc|cccc}
      \toprule
      \multirow{2}{*}{Methods} & \multirow{2}{*}{Modality} &\multicolumn{4}{c|}{\makecell{Query: 2021-04-29-15-55 \\ Database: 2020-12-18-13-44}}&\multicolumn{4}{c|}{\makecell{Query: 2021-09-14-20-00 \\ Database: 2020-12-18-13-44}}&\multicolumn{4}{c}{\makecell{Query: 2021-11-16-14-10 \\ Database: 2020-12-18-13-44}} \\ \cline{3-14}
      ~ & ~ & AR@1 & AR@5 & AR@10 & max $F_{1}$ & AR@1 & AR@5 & AR@10  & max $F_{1}$ & AR@1 & AR@5 & AR@10  & max $F_{1}$ \\ \hline
      AnyLoc~\cite{keetha2023anyloc} & C & 71.15 & 86.23 & 90.43 & 0.8368 & 22.31 & 37.07 & 43.04 & 0.3648 & 77.90 & 91.18 & 94.37 & 0.8781 \\
      MegaLoc~\cite{berton2025megaloc} & C & 95.32 & 96.07 & 96.27 & 0.9848 & 53.64 & 73.79 & 79.95 & 0.7585 & 95.12 & 96.06 & 96.12 & \underline{0.9844} \\
      EgoNN~\cite{komorowski2021egonn} & L & \underline{96.33} & \underline{96.45} & \underline{96.48} & 0.9814 & 94.00 & 95.18 & 95.54 & 0.9741 & \underline{96.03} & \underline{96.30} & \underline{96.33} & 0.9797 \\
      BEVPlace++~\cite{luo2024bevplace++} & L & 84.41 & 89.45 & 91.12 & 0.9272 & 75.46 & 85.53 & 88.02 & 0.8651 & 83.93 & 90.34 & 92.02 & 0.9660 \\
      OpenRadVLAD~\cite{gadd2024open} & R & 53.30 & 68.18 & 74.67 & 0.7129 & 19.26 & 27.95 & 31.91 & 0.5088 & 0.8962 & 2.688 & 4.059 & 0.3200 \\
      LRFusionPR~\cite{qi2025lrfusionpr} & L+R & 96.04 & 96.33 & 96.36 & \underline{0.9851} & \underline{95.24} & \underline{95.60} & \underline{95.67} & \underline{0.9931} & 21.73 & 47.41 & 63.12 & 0.3571 \\
      \rowcolor{gray!25}
      UniMPR (ours) & L+C+R & \textbf{99.40} & \textbf{99.81} & \textbf{99.87} & \textbf{0.9975} & \textbf{99.56} & \textbf{99.83} & \textbf{99.92} & \textbf{0.9979} & \textbf{98.36} & \textbf{99.04}  & \textbf{99.36} & \textbf{0.9931} \\
      \bottomrule
    \end{tabular}
    }
    \label{tab:boreas}
    \vspace{-0.3cm}
\end{table*}

\begin{table*}[ht]
  \centering
  	\vspace{0.0cm}
  	\setlength{\tabcolsep}{4.4pt}
    \setlength{\abovecaptionskip}{0.15cm}
    \caption{Performance under adverse weather conditions}
    \resizebox{\textwidth}{!}{
    \begin{tabular}{c|c|cccc|cccc|cccc}
      \toprule
      \multirow{3}{*}{\makecell{\\ Methods}} & \multirow{3}{*}{\makecell{\\ Modality}} & \multicolumn{4}{c|}{Rain} & \multicolumn{4}{c|}{Simulated Fog} & \multicolumn{4}{c}{Snow} \\ \cline{3-14}
      ~ & ~ & \multicolumn{4}{c|}{\makecell{Query: 2019-01-16-14-15-33 \\ Database: 2019-01-10-14-50-05}} & \multicolumn{4}{c|}{SQ split} & \multicolumn{4}{c}{\makecell{Query: 2021-01-26-11-22 \\ Database: 2020-12-18-13-44}} \\ \cline{3-14}
      ~ & ~ & AR@1 & AR@5 & AR@10 & max $F_{1}$ & AR@1 & AR@5 & AR@10  & max $F_{1}$ & AR@1 & AR@5 & AR@10  & max $F_{1}$ \\ \hline
      AnyLoc~\cite{keetha2023anyloc} & C & 98.13 & 99.31 & 99.50 & 0.9922 & 54.05 & 62.28 & 66.63 & 0.8323 & 72.05 & 88.88 & 92.50 & 0.8459 \\
      MegaLoc~\cite{berton2025megaloc} & C & \underline{99.17} & \underline{99.57} & \underline{99.64} & \underline{0.9975} & 54.05 & 60.52 & 63.92 & 0.8899 & 96.57 & 97.74 & 97.95 & 0.9877 \\
      EgoNN~\cite{komorowski2021egonn} & L & 93.96 & 97.43 & 98.57 & 0.9700 & 49.47 & 62.51 & 66.27 & 0.7635 & 82.46 & 90.77 & 93.07 & 0.9797 \\
      BEVPlace++~\cite{luo2024bevplace++} & L & 96.52 & 98.25 & 98.60 & 0.9824 & 36.08 & 52.76 & 63.92 & 0.5316 & 75.51 & 82.18 & 85.22 & 0.8619 \\
      AutoPlace~\cite{cait2022autoplace} & CR & - & - & - & - & 64.12 & 66.12 & 68.00 & 0.9411 & - & - & - & - \\
      OpenRadVLAD~\cite{gadd2024open} & SR & 86.80 & 91.99 & 94.43 & 0.9334 & - & - & - & - & 93.42 & 97.94 & \underline{99.00} & 0.9674 \\
      LCPR~\cite{zhou2023lcpr} & L+C & - & - & - & - & 13.98 & 27.03 & 37.49 & 0.2872 & - & - & - & - \\
      LRFusionPR~\cite{qi2025lrfusionpr} & L+R & 98.69 & 99.20 & 99.37 & 0.9953 & \underline{91.07} & \underline{96.47} & \underline{97.53} & \underline{0.9596} & \underline{97.84} & \underline{98.04} & 98.09 & \underline{0.9900} \\
      \rowcolor{gray!25}
      UniMPR (ours) & L+C+R & \textbf{99.99} & \textbf{99.99} & \textbf{100.0} & \textbf{0.9999} & \textbf{96.94} & \textbf{99.41} & \textbf{99.88} & \textbf{0.9851} & \textbf{99.70} & \textbf{99.91}  & \textbf{99.98} & \textbf{0.9985} \\
      \bottomrule
    \multicolumn{14}{p{0.9\textwidth}}{C: Camera, L: LiDAR, CR: Single-Chip Radar, SR: Scanning Radar, R: Compatible with both Single-Chip and Scanning Radar.}\\
    \multicolumn{14}{p{0.9\textwidth}}{Missing entries in the table indicate that the session could not meet the input data requirements for the corresponding method.}\\
    \end{tabular}
    }
    \label{tab:weather}
    \vspace{-0.1cm}
\end{table*}

\begin{figure*}[t]
\centerline{\includegraphics[width=\textwidth]{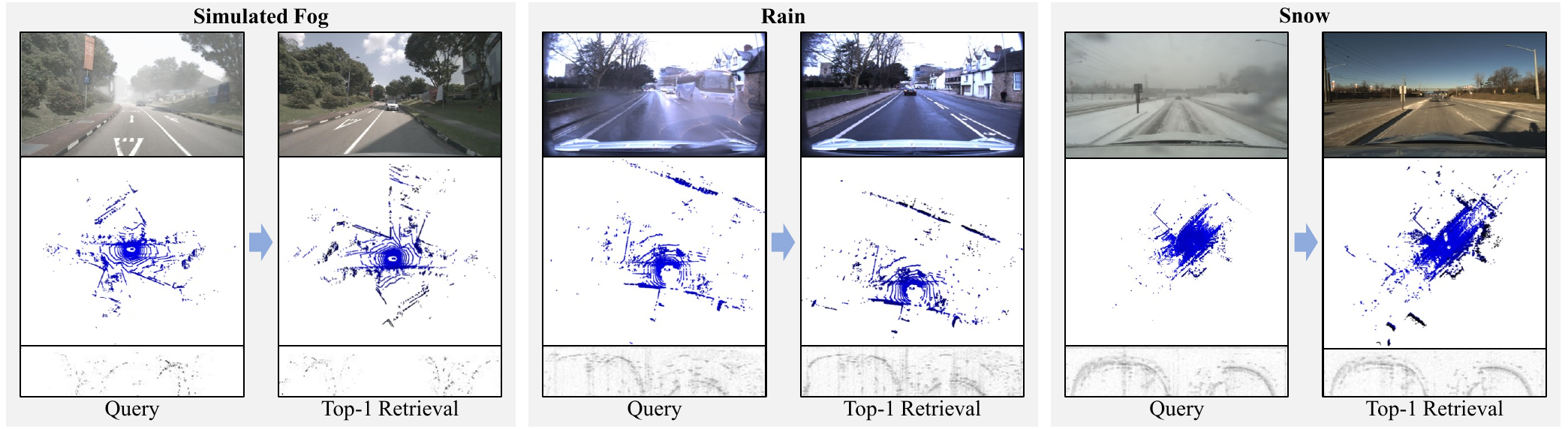}}
\caption{Visualization examples of place retrieval results under different adverse weather conditions.}
\label{fig_weather}
\vspace{-0.4cm}
\end{figure*}

As shown in~\tabref{tab:nclt}, on the NCLT dataset where radar data is absent, UniMPR not only maintains robust performance but significantly outperforms all unimodal and multimodal baselines. Notably, our method maintains consistent performance over a one-year time span, confirming that the large-scale pre-training endows it with the ability to extract sufficiently robust environmental features. Similarly, in camera-absent conditions on the MulRan dataset (see~\tabref{tab:mulran}), our method delivers competitive performance against SOTA MPR method specifically designed for this modality combination. When evaluating on the LiDAR-excluded nuScenes benchmark (see~\tabref{tab:nusc_lidar_exclude}), both LCPR and LRFusionPR exhibit significant performance degradation. In contrast, our method even outperforms the dedicated camera-radar fusion method CRPlace.

These experiments confirm that UniMPR fundamentally transcends the limitations of conventional MPR designs. Rather than limited to fixed modality combinations, our framework maintains high performance across various modality combinations. This flexibility ensures reliable deployment in real-world scenarios where hardware configurations vary across platforms or operational conditions demand adaptive resource utilization.

\subsection{Zero-Shot Generalization Performance}
\label{sec:zero_shot}
In this section, we conduct experiments to validate our model's zero-shot generalization capability across unseen environments and sensor configurations. The evaluation employs KITTI and Boreas datasets that are completely disjoint from the training set in temporal, spatial, and sensor setups, thereby verifying the plug-and-play capability that distinguishes our method from existing MPR solutions.

We perform loop closure detection experiments on the KITTI dataset following OverlapTransformer~\cite{ma2022overlaptransformer}. The results are shown in~\tabref{tab:kitti} and~\figref{fig:kitti_pr}. Due to the limited modality availability in KITTI, which contains only front-facing camera and 64-beam LiDAR data, the table omits baselines like LCPR and LRFusionPR that depend on surround-view images, 32-beam LiDAR or radar data. Nevertheless, our proposed method demonstrates robust performance under these modality constraints, showcasing its generalization capability.

Furthermore, we perform place recognition experiments on the Boreas dataset, which contains front-facing camera, 128-beam LiDAR and scanning radar. In addition to the challenge posed by long-term environmental changes, our Boreas benchmark also includes scenarios with modality degradation. For instance, the sequence pairs ``2021-09-14-20-00'' and ``2020-12-18-13-44'' exhibit a day-night difference, which degrades the performance of unimodal VPR methods. Additionally, the sequence ``2021-11-16-14-10'' is captured using a radar different from the one used for the database, creating a significant domain gap that is difficult for RPR methods to handle. In contrast, our proposed method successfully overcomes these combined challenges, as shown in~\tabref{tab:boreas}. This further substantiates the adaptability and zero-shot generalization capability of our proposed method.

\subsection{Robustness to Adverse Weather Conditions}
In this section, we conduct experiments under diverse adverse weather conditions to further evaluate the performance of our model in the presence of perceptual degradation. To ensure comprehensive evaluation, we conducted experiments under three common adverse weather conditions: rain, fog, and snow. As detailed in~\secref{sec:dataset_setups}, we employ the Oxford Radar and Boreas datasets to construct query sets affected by rain and snow, respectively. Given the lack of highly suitable foggy weather data for place recognition evaluation, we follow~\cite{qi2025lrfusionpr} to generate simulated foggy data on the nuScenes-SQ split. Specifically, the fog model provided by SeeingThroughFog~\cite{bijelic2020seeing} is applied to process both camera and LiDAR data, simulating foggy conditions with a visibility of approximately 50\,meters. ~\figref{fig_weather} depicts a comparison between queries under adverse weather conditions and the top-1 candidate under clear weather.

The results are presented in~\tabref{tab:weather}. Notably, severe weather conditions, especially fog and snow, degrade the performance of unimodal VPR and LPR methods. In contrast, our method maintains robustness across all conditions. This resilience is largely attributed to the fusion of radar's inherent robustness into our multimodal descriptor. Furthermore, our method achieves superior accuracy compared to both RPR methods and radar-fusion MPR method, highlighting the performance gains obtained from fully exploiting intra‑modal features.

\begin{figure}[t]
\centerline{\includegraphics[width=\columnwidth]{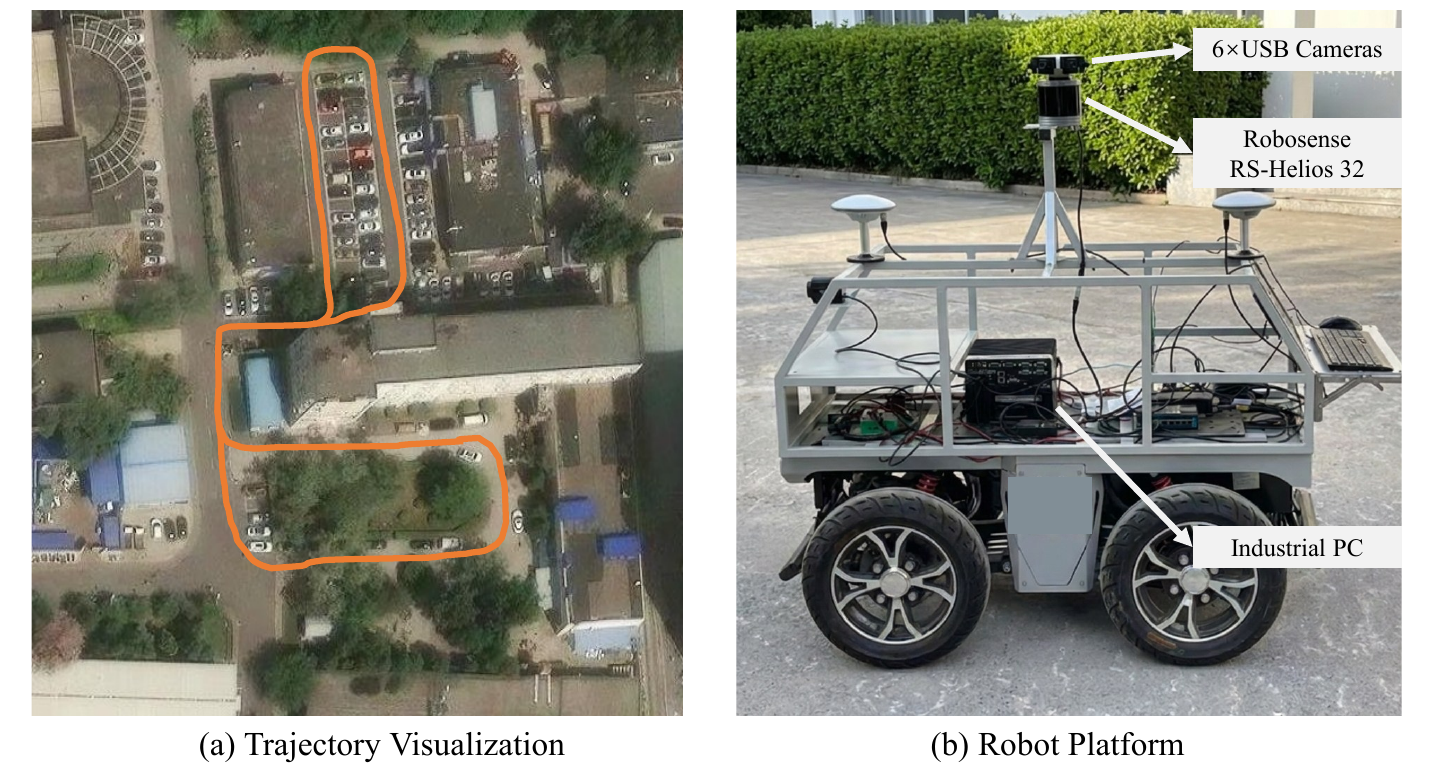}}
\caption{The trajectory visualization and the robot platform of our self-collected dataset.}
\label{fig:self_collected}
\vspace{-0.5cm}
\end{figure}

\begin{table}[t]
  \centering
  \vspace{0.0cm}
  \setlength{\tabcolsep}{4.5pt}
  \caption{Performance on the self-collected dataset}
  \resizebox{\columnwidth}{!}{
    \begin{tabular}{c|c|cccc}
        \toprule
        Methods & Modality & AR@1 & AR@5 & AR@10 & max $F_{1}$ \\ \hline
        AnyLoc~\cite{keetha2023anyloc} & C & 67.26 & 83.63 & 90.04 & 0.7427 \\
        MegaLoc~\cite{berton2025megaloc} & C & 71.17 & 85.41 & 90.39 & 0.7928 \\
        EgoNN~\cite{komorowski2021egonn} & L & 66.67 & 73.16 & 76.40 & \textbf{0.8144} \\
        BEVPlace++~\cite{luo2024bevplace++} & L & \underline{72.60} & \underline{90.39} & \underline{92.17} & 0.7514 \\
        LCPR~\cite{zhou2023lcpr} & L+C & 37.76 & 52.80 & 62.83 & 0.5482 \\
        \rowcolor{gray!25}
        UniMPR$^{*}$ (ours) & L+C & \textbf{80.78} & \textbf{89.68} & \textbf{96.09} & \underline{0.8035} \\
        \bottomrule
    \end{tabular}
  }
  \label{tab:self_coll}
  \vspace{-0.5cm}
\end{table}

\subsection{Real-World Experiments for Place Recognition}
\label{sec:real_world}
In this section, to validate our claim that the proposed method can be deployed without fine-tuning on common computationally capable platforms, we conduct experiments using a self-collected dataset. \figref{fig:self_collected} shows the robot platform equipped with surround-view USB cameras, an industrial computer
with NVIDIA GTX 1080 GPU and a 32-beam LiDAR used in our experiments, along with a trajectory visualization. We use data collected six weeks prior as the database set, and perform place recognition by re‑traversing a similar route six weeks later. Since the trajectory encompasses both parking lots and the main campus thoroughfares, the primary challenges of this scenario lie in the significant structural changes and the frequent presence of dynamic objects.

Since this environment is relatively small, we consider the retrieval to be successful if the estimated position is within 3\,meters of the query location. As shown in~\tabref{tab:self_coll}, the domain gap resulting from an unseen scenario, different sensor configurations, and evaluation criteria that deviate from the training setup all contribute to a performance impact on the evaluated methods. Notably, MPR methods that require more sensor inputs face a larger domain gap than unimodal approaches, leading to more severe performance degradation. In contrast, our proposed UniMPR, while exhibiting comparable adaptability to unimodal methods, harnesses the advantages of multi-sensor input in terms of localization accuracy.

\begin{table}
\centering
\vspace{0.0cm}
\setlength{\tabcolsep}{2pt}
\caption{Ablation study on modality contributions}
\resizebox{\columnwidth}{!}{
  \begin{tabular}{c|c|ccc|ccc}
    \toprule
    \multirow{2}{*}{Methods} & \multirow{2}{*}{Modality} & \multicolumn{3}{c|}{BS split} & \multicolumn{3}{c}{SQ split} \\ \cline{3-8}
    ~ & ~ & AR@1 & AR@5 & AR@10 & AR@1 & AR@5 & AR@10 \\ \hline
    UniMPR-L & L & 98.70 & 99.68 & 99.86 & 99.18 & 99.88 & 99.88 \\
    UniMPR-C & C & 93.88 & 98.70 & 99.46 & 76.50 & 87.19 & 91.30 \\
    UniMPR-R & R & 64.52 & 73.34 & 76.32 & 72.86 & 83.43 & 87.07 \\
    UniMPR-M & L+C+R & 98.92 & 99.73 & 99.81 & 99.06 & 99.88 & 100.0 \\
    \rowcolor{gray!25}
    UniMPR & L+C+R & \textbf{99.13} & \textbf{99.86} & \textbf{99.92} & \textbf{99.65} & \textbf{99.88} & \textbf{100.0} \\
    \bottomrule
  \end{tabular}
}
\label{tab:modality_contribution}
\vspace{-0.3cm}
\end{table}

\begin{table}
\centering
\vspace{0.0cm}
\setlength{\tabcolsep}{4pt}
\caption{Ablation study between polar BEV and Cartesian BEV}
\resizebox{\columnwidth}{!}{
  \begin{tabular}{c|ccc|ccc}
    \toprule
    \multirow{2}{*}{Methods} & \multicolumn{3}{c|}{BS split} & \multicolumn{3}{c}{SQ split} \\ \cline{2-7}
    ~ & AR@1 & AR@5 & AR@10 & AR@1 & AR@5 & AR@10 \\ \hline
    UniMPR-Cart & 90.55 & 93.04 & 94.29 & 81.92 & 91.79 & 95.78 \\
    \rowcolor{gray!25}
    UniMPR & 99.13 & 99.86 & 99.92 & 99.65 & 99.88 & 100.0 \\
    \bottomrule
  \end{tabular}
}
\label{tab:bev_ablation}
\vspace{-0.5cm}
\end{table}

\subsection{Ablation Studies}
\subsubsection{Modality Contributions}
In~\tabref{tab:modality_contribution}, we present an ablation study on the contribution of each modality to the overall place recognition performance. We denote the three modality-specific branches as UniMPR-L, UniMPR-C, and UniMPR-R, and the fusion branch as UniMPR-M. 
Notably, even when individually evaluated, the modality-specific branches achieve competitive performance compared to unimodal baselines (see~\tabref{tab:nuscenes}). The fusion branch (UniMPR-M) is designed to exploit cross-modal feature correlations, handling scenarios where the modality-specific branches fall short. Nevertheless, UniMPR-M alone does not yield optimal results, likely due to the loss of modality-specific information during fusion. By concatenating both modality-specific and fusion descriptors, the model fully leverages multimodal information and attains the best performance.

\subsubsection{Polar BEV Representation}
\label{sec:polar_bev_representation}
In this section, we conduct ablation studies to validate the rationale of adopting the polar BEV as the unified representation.

Firstly, we demonstrate that using the polar BEV yields better recognition accuracy compared to Cartesian BEV. We modify the representation of all three modalities to Cartesian BEV, denoting this variant as UniMPR-Cart. \tabref{tab:bev_ablation} show that UniMPR with polar BEV outperforms UniMPR-Cart across all metrics. This phenomenon is likely attributed to the robustness of polar BEV to viewpoint changes, especially rotation variations.

Subsequently, we test the robustness of both polar BEV and Cartesian BEV representations to viewpoint changes. We apply random rotations of $(0, 2\pi)$, random translations within $(-4.5, 4.5)$ meters, and a combination of both to the query set. Due to the difficulty in simulating the corresponding changes in the images, we conduct these experiments on the MulRan-Sejong. As shown in~\tabref{tab:viewpoint_change}, the model exhibits greater robustness to viewpoint variations when using polar BEV as the unified representation. This observation corroborates the hypothesis derived from~\tabref{tab:bev_ablation}. The enhancement of rotation invariance stems from the inherent property of polar BEV, which transforms spatial rotations into column-wise cyclic shifts. In contrast to Cartesian BEV, where rotations result in complex non-linear transformations, this property of polar BEV allows the network to more easily extract rotation-invariant features. Notably, the polar BEV also retains strong translation robustness. We attribute this to the highly compact representation of the feature map. Although translation induces non‑linear pixel displacement in polar coordinates, the scale of these transformations is insufficient to significantly degrade the recognition performance.

We also ablate the use of point density as the value for polar BEV. We compare it with a variant that uses maximum elevation as the value, similar to Scan Context~\cite{kim2018scan}. We conduct experiments using the LiDAR-specific branch, UniMPR-L, to highlight the impact of the polar BEV value. \tabref{tab:value_ablation} indicates that using maximum elevation may fail to suppress noise effectively in the same way as point density, consequently impairing its recognition performance.

\begin{table}[t!]
\centering
\vspace{0.0cm}
\setlength{\tabcolsep}{2.5pt}
\caption{Robustness to viewpoint variations}
\resizebox{\columnwidth}{!}{
  \begin{tabular}{c|c|ccc}
    \toprule
    \multirow{2}{*}{Methods} & \multirow{2}{*}{\makecell{Viewpoint \\ Changes}} & \multicolumn{3}{c}{Sejong} \\ \cline{3-5}
    ~ & ~ & AR@1 & AR@5 & AR@10 \\ \hline
    \multirow{4}{*}{UniMPR-Cart} & Identity & 97.17 & 99.49 & 99.93 \\
    ~ & Rot & 38.55 (\textcolor{red}{-58.62}) & 56.74 (\textcolor{red}{-42.75}) & 64.57 (\textcolor{red}{-35.36}) \\
    ~ & Trans & 93.72 (\textcolor{red}{-3.45}) & 96.30 (\textcolor{red}{-3.19}) & 97.04 (\textcolor{red}{-2.89}) \\
    ~ & Rot+Trans & 23.91 (\textcolor{red}{-73.26}) & 37.10 (\textcolor{red}{-62.39}) & 42.61 (\textcolor{red}{-57.32}) \\ \hline
    \multirow{4}{*}{UniMPR} & Identity & 98.62 & 99.78 & 100.0 \\
    ~ & Rot & 95.80 (\textcolor{red}{-2.82}) & 98.62 (\textcolor{red}{-1.16}) & 99.28 (\textcolor{red}{-0.72}) \\
    ~ & Trans & 97.17(\textcolor{red}{-1.45}) & 99.42 (\textcolor{red}{-0.36}) & 99.71 (\textcolor{red}{-0.29}) \\
    ~ & Rot+Trans & 94.49 (\textcolor{red}{-4.13}) & 98.04 (\textcolor{red}{-1.74}) & 98.77 (\textcolor{red}{-1.23}) \\
    \bottomrule
  \end{tabular}
}
\label{tab:viewpoint_change}
\vspace{-0.3cm}
\end{table}

\begin{table}[t]
\centering
\vspace{0.0cm}
\setlength{\tabcolsep}{4pt}
\caption{Ablation study on polar BEV value representations}
\resizebox{\columnwidth}{!}{
  \begin{tabular}{c|ccc|ccc}
    \toprule
    \multirow{2}{*}{Methods} & \multicolumn{3}{c|}{BS split} & \multicolumn{3}{c}{SQ split} \\ \cline{2-7}
    ~ & AR@1 & AR@5 & AR@10 & AR@1 & AR@5 & AR@10 \\ \hline
    Max Elevation & 97.84 & 98.94 & 99.11 & 96.51 & 98.62 & 98.92 \\
    \rowcolor{gray!25}
    Point Density & \textbf{98.70} & \textbf{99.68} & \textbf{99.86} & \textbf{99.18} & \textbf{99.88} & \textbf{99.88} \\
    \bottomrule
  \end{tabular}
}
\label{tab:value_ablation}
\vspace{-0.3cm}
\end{table}

\begin{table}[t!]
\centering
\vspace{0.0cm}
\setlength{\tabcolsep}{4.6pt}
\caption{Ablation study on different image backbones}
\resizebox{\columnwidth}{!}{
  \begin{tabular}{c|ccc|ccc}
    \toprule
    \multirow{2}{*}{Backbones} & \multicolumn{3}{c|}{BS split} & \multicolumn{3}{c}{Boreas} \\ \cline{2-7}
    ~ & AR@1 & AR@5 & AR@10 & AR@1 & AR@5 & AR@10 \\ \hline
    ResNet18 & 80.28 & 90.37 & 92.53 & 44.44 & 59.14 & 65.01\\
    DINOv2 & 92.32 & 97.16 & 98.46 & 67.24 & 82.30 & 87.12 \\
    \rowcolor{gray!25}
    DINOv3 & \textbf{93.88} & \textbf{98.70} & \textbf{99.46} & \textbf{83.16} & \textbf{90.58} & \textbf{92.46} \\
    \bottomrule
  \end{tabular}
}
\label{tab:backbone_ablation}
\vspace{-0.5cm}
\end{table}

\subsubsection{Image Backbone}
\label{sec:image_backbone}
In this section, we conduct an ablation study on the choice of the image backbone. We replace DINOv3 with ResNet18 and DINOv2~\cite{oquab2023dinov2}, respectively, and evaluate the performance of UniMPR-C. As shown in~\tabref{tab:backbone_ablation}, even on the BS split, which has relatively high similarity to the training set, the performance of ResNet18 lags behind that of the well pre-trained DINOv2 and DINOv3. Furthermore, in zero-shot generalization experiments on completely unseen datasets such as Boreas ``2021-04-29-15-55''-``2020-12-18-13-44'', using DINOv3 as the backbone shows superior performance compared to DINOv2. This demonstrates the significance of the domain-invariant features extracted by DINOv3-S for the generalization capability of the camera branch UniMPR-C.

\begin{table}
\centering
\vspace{0.0cm}
\setlength{\tabcolsep}{5.1pt}
\caption{Ablation study on the Gaussian BEV projection module}
\resizebox{\columnwidth}{!}{
  \begin{tabular}{c|c|c|ccc}
    \toprule
    \multirow{2}{*}{Methods} & \multirow{2}{*}{\makecell{Inference \\ Time [ms]}} & \multirow{2}{*}{GFLOPs} & \multicolumn{3}{c}{BS split} \\ \cline{4-6}
    ~ & ~ & ~ & AR@1 & AR@5 & AR@10 \\ \hline
    UniMPR-C-LSS & 16.85 & 20.03 & 87.45 & 92.51 & 93.56 \\
    UniMPR-C-S & 12.34 & 19.77 & 93.59 & 98.24 & 99.11 \\
    UniMPR-C-SR & 12.91 & 19.78 & 91.91 & 98.40 & 99.49 \\
    \rowcolor{gray!25}
    UniMPR & \textbf{12.24} & \textbf{19.76} & \textbf{93.88} & \textbf{98.70} & \textbf{99.46} \\
    \bottomrule
  \end{tabular}
}
\label{tab:lifting_ablation}
\vspace{-0.3cm}
\end{table}

\subsubsection{Gaussian BEV Projection Module}
\label{sec:gaussian_bev_projection_module}
In this section, we demonstrate the design rationale of the proposed Gaussian BEV projection module through ablation studies. Moreover, we showcase the basis for treating the scale and rotation attributes of the Gaussians as constants.

Ablation experiments are performed and reported based on the unimodal camera branch, UniMPR‑C. First, we replace the Gaussian BEV projection module with Lift‑Splat‑Shoot~\cite{philion2020lift}, a representative BEV‑lifting approach, and denote this variant as UniMPR‑C‑LSS. For fair comparison, the aligned DINOv3 features are fed into the LSS module. Furthermore, the variant that predicts scale with an MLP is denoted as UniMPR‑C‑S, and the variant that uses two separate MLPs to predict scale and rotation is denoted as UniMPR‑C‑SR. As shown in~\tabref{tab:lifting_ablation}, although UniMPR‑C‑LSS requires more computation and longer inference time, its performance still falls short compared to UniMPR‑C. This may be because, for place recognition tasks, how to encode image features reasonably in the BEV feature map is more critical than recovering the spatial structure of the scene. The fact that UniMPR‑C‑S and UniMPR‑C‑SR also fail to outperform UniMPR‑C further suggests this point.

\begin{table}
\centering
\vspace{0.0cm}
\setlength{\tabcolsep}{4.5pt}
\caption{Ablation study on MoE Transformer}
\resizebox{\columnwidth}{!}{
  \begin{tabular}{c|ccc|ccc}
    \toprule
    \multirow{2}{*}{\makecell{MoE \\ Transformer}} & \multicolumn{3}{c|}{BS split} & \multicolumn{3}{c}{SQ split} \\ \cline{2-7}
    ~ & AR@1 & AR@5 & AR@10 & AR@1 & AR@5 & AR@10 \\ \hline
    \ding{55} & 98.40 & 99.40 & 99.62 & 97.88 & 99.76 & 100.0 \\
    \rowcolor{gray!25}
    \ding{51} & \textbf{98.92} & \textbf{99.73} & \textbf{99.81} & \textbf{99.06} & \textbf{99.88} & \textbf{100.0} \\
    \bottomrule
  \end{tabular}
}
\label{tab:moe_transformer}
\vspace{-0.5cm}
\end{table}

\subsubsection{MoE Transformer}
\label{sec:moe_transformer}
In this section, we perform an ablation study to examine the rationale of utilizing the MoE Transformer in multimodal fusion. Specifically, we replace the MoE Transformer in the fusion branch UniMPR‑M with a standard transformer encoder. As shown in~\tabref{tab:moe_transformer}, this modification leads to a degradation in recognition accuracy. This suggests that, the MoE Transformer mitigates gradient conflicts between different modalities during optimization through its sparsely-activated expert mechanism, thereby enabling the model to converge to a more favorable optimum. This, in turn, enhances the recognition accuracy.

\subsubsection{Feature Aggregator}
\label{sec:feature_aggregator}
In this section, we conduct an ablation to evaluate the effectiveness of NetVLAD as the feature aggregator. We compare NetVLAD with global average pooling (GAP), global max pooling (GMP), and GeM pooling. As shown in~\tabref{tab:aggregator}, employing NetVLAD as the feature aggregator achieves better recognition accuracy. This may be because, in general MPR tasks, which are affected by viewpoint changes and various sensor configurations, extracting higher-level abstract representations is more effective than preserving local structures.

\begin{table}
\centering
\vspace{0.0cm}
\setlength{\tabcolsep}{4.3pt}
\caption{Ablation study on different feature aggregators}
\resizebox{\columnwidth}{!}{
  \begin{tabular}{c|ccc|ccc}
    \toprule
    \multirow{2}{*}{Aggregators} & \multicolumn{3}{c|}{BS split} & \multicolumn{3}{c}{SQ split} \\ \cline{2-7}
    ~ & AR@1 & AR@5 & AR@10 & AR@1 & AR@5 & AR@10 \\ \hline
    GAP & 95.56 & 98.11 & 98.57 & 90.25 & 97.18 & 98.59 \\
    GMP & 93.72 & 97.46 & 98.29 & 87.43 & 94.01 & 96.36 \\
    GeM & 94.18 & 97.32 & 98.46 & 85.90 & 95.42 & 98.59 \\
    \rowcolor{gray!25}
    NetVLAD & \textbf{99.13} & \textbf{99.86} & \textbf{99.92} & \textbf{99.65} & \textbf{99.88} & \textbf{100.0} \\
    \bottomrule
  \end{tabular}
}
\label{tab:aggregator}
\vspace{-0.3cm}
\end{table}

\begin{table}
\centering
\vspace{0.0cm}
\setlength{\tabcolsep}{6pt}
\caption{Ablation study on learnable BEV imputation}
  \begin{tabular}{c|ccc|ccc}
    \toprule
    \multirow{2}{*}{LBI} & \multicolumn{3}{c|}{Sejong} & \multicolumn{3}{c}{KITTI 00} \\ \cline{2-7}
    ~ & AR@1 & AR@5 & AR@10 & AR@1 & AR@5 & AR@10 \\ \hline
    \ding{55} & 98.48 & 99.78 & 99.93 & 95.82 & 97.31 & 97.61 \\
    \rowcolor{gray!25}
    \ding{51} & \textbf{98.62} & \textbf{99.78} & \textbf{100.00} & \textbf{97.31} & \textbf{97.61} & \textbf{97.61} \\
    \bottomrule
  \end{tabular}
\label{tab:learnable_bev_imputation}
\vspace{-0.5cm}
\end{table}

\subsubsection{Learnable BEV Imputation}
\label{sec:learnable_bev_imputation}
In this section, we conduct an ablation study to evaluate the role of the learnable BEV imputation (LBI). We compare UniMPR with a variant that simply pads missing modalities with zero vectors. As shown in~\tabref{tab:learnable_bev_imputation}, on the MulRan dataset (no camera) and the KITTI dataset (no radar), using the learnable BEV imputation leads to consistent performance improvements. By using learnable weights for feature imputation, UniMPR minimizes the impact of missing modalities on exploiting cross‑modal feature correlations, thereby handling various modality inputs more effectively.

\begin{table}[t]
\centering
\vspace{0.0cm}
\setlength{\tabcolsep}{2.3pt}
\caption{Ablation study on two-stage training strategy}
\resizebox{\columnwidth}{!}{
  \begin{tabular}{c|c|ccc|ccc}
    \toprule
    \multirow{2}{*}{Methods} & \multirow{2}{*}{TST} & \multicolumn{3}{c|}{BS split} & \multicolumn{3}{c}{SQ split} \\ \cline{3-8}
    ~ & ~ & AR@1 & AR@5 & AR@10 & AR@1 & AR@5 & AR@10 \\ \hline
    UniMPR-NU-L & \multirow{5}{*}{\ding{55}} & 95.37 & 97.54 & 98.29 & 96.59 & 98.71 & 99.18 \\
    UniMPR-NU-C & ~ & 86.22 & 95.51 & 97.46 & 73.21 & \textbf{88.95} & \textbf{92.24} \\
    UniMPR-NU-R & ~ & 46.96 & 60.87 & 67.17 & 66.51 & 79.20 & 84.02 \\
    UniMPR-NU-M & ~ & 93.83 & 96.64 & 97.59 & 95.30 & 98.35 & 98.82 \\
    UniMPR-NU & ~ & 98.48 & 99.54 & 99.73 & \textbf{98.82} & 99.76 & 99.88 \\ \cline{1-8}
    UniMPR-NU-L & \multirow{5}{*}{\ding{51}} & \textbf{98.00} & \textbf{99.30} & \textbf{99.57} & \textbf{97.30} & \textbf{99.53} & \textbf{99.76} \\
    UniMPR-NU-C & ~ & \textbf{90.15} & \textbf{97.54} & \textbf{98.51} & \textbf{75.68} & 87.54 & 89.66 \\
    UniMPR-NU-R & ~ & \textbf{69.21} & \textbf{76.52} & \textbf{79.25} & \textbf{74.03} & \textbf{84.14} & \textbf{88.84} \\
    UniMPR-NU-M & ~ & \textbf{98.24} & \textbf{99.76} & \textbf{99.81} & \textbf{96.83} & \textbf{99.76} & \textbf{99.88} \\
    UniMPR-NU & ~ & \textbf{98.76} & \textbf{99.81} & \textbf{99.92} & 98.12 & \textbf{99.88} & \textbf{100.0} \\
    \bottomrule
  \end{tabular}
}
\label{tab:training_strategy}
\vspace{-0.3cm}
\end{table}

\begin{table}[t]
\centering
\vspace{0.0cm}
\setlength{\tabcolsep}{2.3pt}
\caption{Ablation study on two-stage training strategy for LiDAR modality missing scenarios}
\resizebox{\columnwidth}{!}{
  \begin{tabular}{c|c|ccc|ccc}
    \toprule
    \multirow{2}{*}{Methods} & \multirow{2}{*}{TST} & \multicolumn{3}{c|}{BS split} & \multicolumn{3}{c}{SQ split} \\ \cline{3-8}
    ~ & ~ & AR@1 & AR@5 & AR@10 & AR@1 & AR@5 & AR@10 \\ \hline
    UniMPR-NU & \ding{55} & 75.99 & 86.82 & 90.72 & 77.09 & 92.48 & 95.30 \\ 
    UniMPR-NU & \ding{51} & \textbf{88.66} & \textbf{95.21} & \textbf{96.37} & \textbf{86.13} & \textbf{95.65} & \textbf{97.77} \\
    \bottomrule
  \end{tabular}
}
\label{tab:training_strategy_missing_modality}
\vspace{-0.3cm}
\end{table}

\begin{table}[t!]
\centering
\vspace{0.0cm}
\setlength{\tabcolsep}{0.8pt}
\caption{Ablation study on dataset and label assessment strategy}
\resizebox{\columnwidth}{!}{
  \begin{tabular}{c|ccc|cc|cc|cc}
    \toprule
    \multirow{2}{*}{Methods} & \multirow{2}{*}{TST} & \multirow{2}{*}{UMD} & \multirow{2}{*}{ALA} & \multicolumn{2}{c|}{SQ} & \multicolumn{2}{c|}{Snow} & \multicolumn{2}{c}{02} \\ \cline{5-10}
    ~ & ~ & ~ & ~ & AR@1 & AR@5 & AR@1 & AR@5 & AR@1 & AR@5 \\ \hline
    UniMPR-NU & \ding{55} & \ding{55} & \ding{55} & 98.82 & 99.76 & 94.80 & 97.93 & 92.54 & 95.22 \\ 
    UniMPR-NU & \ding{51} & \ding{55} & \ding{55} & 98.12 & 99.88 & 87.20 & 90.40 & 95.82 & 96.42 \\
    UniMPR & \ding{51} & \ding{51} & \ding{55} & 98.71 & 99.88 & 99.19 & 99.77 & 97.01 & \textbf{98.51} \\
    \rowcolor{gray!25}
    UniMPR & \ding{51} & \ding{51} & \ding{51} & \textbf{99.65} & \textbf{99.88} & \textbf{99.70} & \textbf{99.91} & \textbf{97.31} & 97.61 \\
    \bottomrule
  \end{tabular}
}
\label{tab:dataset_label_strategy}
\vspace{-0.5cm}
\end{table}

\subsubsection{Training Strategies}
\label{sec:training_strategies}
In this section, we conduct an ablation study on our key training strategies, which includes the two-stage training pipeline (TST), the unified multimodal dataset (UMD), and the adaptive label assessment strategy (ALA). We aim to demonstrate their role in handling various modality inputs, enhancing robustness to missing modalities, and enabling generalization across sensor configurations.

We first denote the model trained solely on the nuScenes dataset as UniMPR‑NU. We train UniMPR‑NU using both a single‑stage end‑to‑end approach and our proposed two‑stage training strategy. As shown in~\tabref{tab:training_strategy}, although end‑to‑end training can theoretically achieve more coordinated global optimization, the two‑stage strategy still yields higher accuracy. This improvement is likely attributable to the ability of the two‑stage strategy to exploit intra‑modal correlations. The support for this interpretation comes from the significant performance gap observed among the unimodal branches (UniMPR‑L, UniMPR‑C, and UniMPR‑R). The capacity of TST to effectively leverage the strengths of individual modalities is crucial for handling various modality inputs and missing modalities. As shown in~\tabref{tab:training_strategy_missing_modality}, when the LiDAR input is missing, the performance of the end‑to‑end variant suffers a drastic drop, while the model trained with the TST strategy maintains more stable performance.

However, due to the limited scale of nuScenes, all variants of UniMPR‑NU show promising but ultimately suboptimal zero‑shot generalization to unseen environments, modalities, and sensor setups. As evidenced in~\tabref{tab:dataset_label_strategy}, their performance remains unsatisfactory on sequences such as the snowy Boreas and KITTI 02. In contrast, training on the unified multimodal dataset enables the model to learn general MPR capabilities that transcend fixed modality combinations and specific sensor configurations, thereby achieving significantly better zero-shot generalization performance. Incorporating the ALA strategy further ensures consistent and effective supervision across all modalities, leading to optimal performance.

\subsection{Sensitivity Analysis}
\subsubsection{Retrieval Threshold}
\begin{figure}[t]
\centerline{\includegraphics[width=\columnwidth]{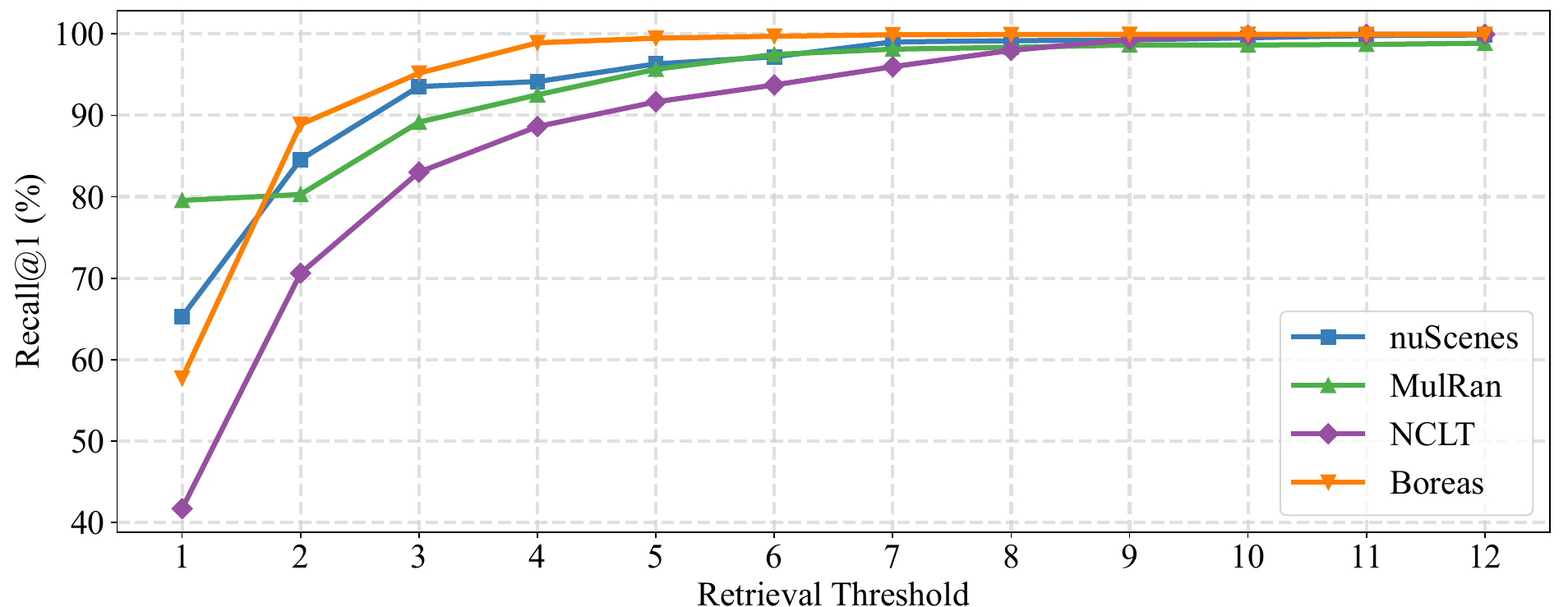}}
\caption{The curves illustrating the Recall@1 change with respect to the success retrieval threshold on various datasets.}
\label{fig:localization_curve}
\vspace{-0.2cm}
\end{figure}

In this section, we conduct a sensitivity analysis on the success retrieval threshold to validate our method's robustness to various settings. We vary the retrieval threshold from 1\,meter to 12\,meters and report the Recall@1 for each case. Experiments are performed on multiple datasets, including the nuScenes-BS split, the MulRan-Sejong split, the sequence pair from ``2012-06-15" to ``2012-01-08" on the NCLT dataset, and the sequence pair from ``2021-04-29-15-55" to ``2020-12-18-13-44" on the Boreas dataset. As shown in~\figref{fig:localization_curve}, our method maintains a high retrieval success rate even when evaluated with a threshold smaller than the $d_{\text{pos}}=9\,\text{m}$. This demonstrates that our model is not merely optimized for a specific setting, but has learned a general capability for similarity-based place recognition by effectively utilizing the similarity between samples. It should be noted, however, that performance declines to some extent when the threshold is set extremely small (e.g., $< 2\,\text{m}$). For applications requiring stricter localization error tolerances, we recommend adjusting the $d_{\text{pos}}$ and $d_{\text{non\_neg}}$, and performing fine-tuning on the pre-trained model.

\begin{table}[t]
\centering
\vspace{0.0cm}
\setlength{\tabcolsep}{10pt}
\caption{Ablation study on two-stage training strategy for LiDAR modality missing scenarios}
\resizebox{\columnwidth}{!}{
  \begin{tabular}{c|cccc}
    \toprule
    \multirow{2}{*}{Noise Level $\tau$} & \multicolumn{4}{c}{BS split} \\ \cline{2-5}
    ~ & AR@1 & AR@5 & AR@10 & max $F_{1}$ \\ \hline
    0 & 99.13 & 99.86 & 99.92 & 0.9959 \\
    1 & 99.08 & 99.80 & 99.84 & 0.9954 \\
    2 & 99.05 & 99.80 & 99.84 & 0.9953 \\
    3 & 99.00 & 99.80 & 99.84 & 0.9950 \\
    4 & 99.02 & 99.80 & 99.84 & 0.9951 \\
    \bottomrule
  \end{tabular}
}
\label{tab:calibration_mismatch}
\vspace{-0.5cm}
\end{table}

\subsubsection{Calibration Noises}
UniMPR utilizes the camera’s intrinsic and extrinsic parameters, along with the calibrated transformations between sensors, to generate the multimodal descriptor. However, in real‑world applications, various factors often make it difficult to obtain perfectly accurate calibration. Therefore, we analyze the sensitivity of UniMPR to calibration mismatch. Following BEVFormer~\cite{li2024bevformer}, we inject rotational noise $N(0, \tau)$ (degrees) and translational noise $N(0, 5\tau)$ (centimeters) into the extrinsic parameters, and varying the noise level $\tau$ from 0 to 4. As shown in~\tabref{tab:calibration_mismatch}, even under the highest noise level, UniMPR does not exhibit significant performance degradation. This robustness likely stems from the relatively coarse resolution of the feature maps In our configuration, to enhance real-time performance, the image features have a resolution of only $21 \times 21$, while the polar BEV feature maps for all modalities are $29 \times 7$. This mitigates the impact of fine-grained calibration misalignments, allowing the model to maintain stable performance.

\subsection{Runtime Efficiency}
In this section, we analyze the runtime efficiency of UniMPR, in terms of inference speed, computational cost, and model parameters. Since the efficiency of UniMPR is directly correlated with the number of camera inputs, we conduct experiments on nuScenes (6 cameras) and Boreas (1 camera), respectively. We report the time to generate a descriptor (extraction time) and to retrieve the top-1 candidate (retrieval time). For a fair comparison, all methods employ the FAISS library~\cite{douze2024faiss} to accelerate retrieval.

As shown in~\tabref{tab:efficiency}, despite fusing multiple modalities, UniMPR’s carefully designed architecture enables real‑time performance with low computational overhead. On the nuScenes dataset, UniMPR achieves better real‑time performance than other foundation model-based methods (e.g, AnyLoc), while remaining competitive with other baselines. Furthermore, on the Boreas dataset, the time required for a single place retrieval further drops to approximately 10\,ms, which is shorter than the refresh rates of most sensors. In terms of computational complexity, UniMPR is comparable to the efficiency-focused baselines like AutoPlace, BEVPlace++ and LRFusionPR, and its 65.8\,M parameters yield a practical memory footprint. These results confirm that UniMPR achieves high accuracy while meeting the computational demands of real‑world deployment on computationally capable autonomous vehicles and robots.

\begin{table}[t]
  \centering
  \setlength{\tabcolsep}{4pt}
  \renewcommand\arraystretch{0.95}
  \setlength{\abovecaptionskip}{0.15cm}
  \caption{Runtime efficiency analysis and comparison}
  \resizebox{\columnwidth}{!}{
  \footnotesize{
    \begin{tabular}{c|c|c|c|c|c}
        \toprule
        Datasets & Methods & Extraction & Retrieval & Total & GFLOPs \\ \hline
        \multirow{7}{*}{nuScenes} & 
        AnyLoc~\cite{keetha2023anyloc} & 87.36 & 49.97 & 137.33 & 314.08 \\
        ~ & MegaLoc~\cite{berton2025megaloc} & 11.19 & 8.52 & 19.71 & 24.08 \\
        ~ & EgoNN~\cite{komorowski2021egonn} & 9.41 & \textbf{0.09} & 9.50 & - \\
        ~ & BEVPlace++~\cite{luo2024bevplace++} & 17.99 & 8.43 & 26.42 & 10.84 \\
        ~ & AutoPlace~\cite{cait2022autoplace} & 21.08 & 4.22 & 25.30 & 20.77 \\
        ~ & LCPR~\cite{zhou2023lcpr} & 8.38 & 0.12 & 8.50 & 57.85 \\
        ~ & LRFusionPR~\cite{qi2025lrfusionpr} & \textbf{6.53} & 0.25 & \textbf{6.78} & \textbf{7.22} \\
        ~ & UniMPR (ours) & 16.45 & 0.52 & 16.97 & 26.91 \\ \hline
        \multirow{7}{*}{Boreas} & 
        AnyLoc~\cite{keetha2023anyloc} & 85.24 & 26.88 & 112.12 & 314.08 \\
        ~ & MegaLoc~\cite{berton2025megaloc} & 10.82 & 4.40 & 15.22 & 22.35 \\
        ~ & EgoNN~\cite{komorowski2021egonn} & 14.00 & \textbf{0.07} & 14.07 & - \\
        ~ & BEVPlace++~\cite{luo2024bevplace++} & 17.53 & 4.31 & 21.84 & 11.49 \\
        ~ & OpenRadVLAD~\cite{gadd2024open} & 51.96 & 47.28 & 99.24 & - \\
        ~ & LRFusionPR~\cite{qi2025lrfusionpr} & \textbf{6.43} & 0.14 & \textbf{6.57} & \textbf{7.54} \\
        ~ & UniMPR (ours) & 9.83 & 0.17 & 10.00 & 14.56 \\
        \bottomrule
    \multicolumn{6}{p{1.0\columnwidth}}{All time measurements are expressed in milliseconds. GFLOPS for sparse convolution-based baselines are not reported due to unmeasurable complexity}\\
    \end{tabular}
    }
  }
  \label{tab:efficiency}
  \vspace{-0.5cm}
\end{table}

\section{Conclusion}
This paper presents a unified MPR framework that can handle various modality inputs, generalize to unseen environments, and adapt to various sensor configurations with a single pre-trained model. This work introduces a multi-branch network architecture, complemented by large-scale pre-training and a stage-wise training scheme. The proposed method aims to overcome the limitations of conventional MPR approaches, such as their restriction to fixed modality combinations, poor generalization to different sensor configurations, and limited robustness to missing or degraded modalities. Extensive experiments on six public datasets and one self-collected dataset demonstrate that the proposed method achieves SOTA performance under diverse and challenging conditions. The ablation studies and comparative analysis further validate the design rationale of each component. In future work, we will focus on distilling this general MPR capability into more compact models and investigating its viability on highly resource-constrained edge devices.

\end{document}